\documentclass{article} 
\usepackage{iclr2024_conference,times}

\iclrfinalcopy

\usepackage{amsmath,amsfonts,bm}

\def\eqref#1{equation~\ref{#1}}

\def\1{\bm{1}}

\def\ra{{\textnormal{a}}}

\def\ro{{\textnormal{o}}}

\def\rr{{\textnormal{r}}}

\def\rva{{\mathbf{a}}}

\def\rvo{{\mathbf{o}}}

\def\va{{\bm{a}}}

\def\vo{{\bm{o}}}

\DeclareMathAlphabet{\mathsfit}{\encodingdefault}{\sfdefault}{m}{sl}
\SetMathAlphabet{\mathsfit}{bold}{\encodingdefault}{\sfdefault}{bx}{n}

\DeclareMathOperator*{\argmax}{arg\,max}
\DeclareMathOperator*{\argmin}{arg\,min}

\newcommand{\te}[1]{\texttt{#1}}

\usepackage{hyperref}
\usepackage{url}

\usepackage{graphicx}
\usepackage{amsmath}
\usepackage{amssymb}
\usepackage{duckuments}
\usepackage{multirow}
\usepackage{tabularx,booktabs}
\usepackage{pifont}
\usepackage{algorithm}
\usepackage{algorithmic}
\usepackage{wrapfig}
\usepackage{enumitem}
\usepackage{subcaption}
\usepackage[page, header]{appendix}
\usepackage{colortbl}
\usepackage{xcolor}
\usepackage[T1]{fontenc}

\title{Learning Multi-Agent Communication from \\ Graph Modeling Perspective}

\author{Shengchao Hu\textsuperscript{1,2},\  Li Shen\textsuperscript{3}\thanks{Corresponding author: Li Shen},\  Ya Zhang\textsuperscript{1,2},\   Dacheng Tao\textsuperscript{4} \\
  \textsuperscript{1} Shanghai Jiao Tong University, \textsuperscript{2} Shanghai AI Laboratory \\
  \textsuperscript{3} JD Explore Academy, 
  \textsuperscript{4} Nanyang Technological University \\
  {\tt\small \{charles-hu,ya\_zhang\}@sjtu.edu.cn; \quad 
\{mathshenli,dacheng.tao\}@gmail.com}
}

\begin{document}

\maketitle

\begin{abstract}
    In numerous artificial intelligence applications, the collaborative efforts of multiple intelligent agents are imperative for the successful attainment of target objectives.
    To enhance coordination among these agents, a distributed communication framework is often employed.
    However, information sharing among all agents proves to be resource-intensive, while the adoption of a manually pre-defined communication architecture imposes limitations on inter-agent communication, thereby constraining the potential for collaborative efforts.
    In this study, we introduce a novel approach wherein we conceptualize the communication architecture among agents as a learnable graph. 
    We formulate this problem as the task of determining the communication graph while enabling the architecture parameters to update normally, thus necessitating a bi-level optimization process. 
    Utilizing continuous relaxation of the graph representation and incorporating attention units, our proposed approach, CommFormer, efficiently optimizes the communication graph and concurrently refines architectural parameters through gradient descent in an end-to-end manner.
    Extensive experiments on a variety of cooperative tasks substantiate the robustness of our model across diverse cooperative scenarios, where agents are able to develop more coordinated and sophisticated strategies regardless of changes in the number of agents.
\end{abstract}
\section{Introduction}

Multi-agent reinforcement learning (MARL) algorithms play an essential role in solving complex decision-making tasks through the analysis of interaction data between computerized agents and simulated or physical environments.
This paradigm finds prevalent application across domains, including autonomous driving \citep{zhou2020smarts, hu2022st}, order dispatching \citep{li2019efficient, yang2018mean}, and gaming AI systems \citep{peng2017multiagent, zhou2023malib}.
In the MARL scenarios typically explored in these studies, multiple agents engage in iterative interactions within a shared environment, continually refining their policies through learning from observations to collectively attain a common objective.
This problem can be conceptually simplified as an instance of independent RL, wherein each agent regards other agents as elements of its environment. 
However, the strategies employed by other agents exhibit dynamic uncertainty and evolve throughout the training process, rendering the environment intrinsically unstable from the viewpoint of each individual agent. 
Consequently, effective collaboration among agents becomes a formidable challenge. 
Additionally, it's important to note that policies acquired through independent RL are susceptible to overfitting with respect to the policies of other agents, as evidenced by \citet{lanctot2017unified}.

Communication is a fundamental pillar in addressing this challenge, serving as a cornerstone of intelligence by enabling agents to operate cohesively as a collective entity rather than disparate individuals.
Its significance becomes especially apparent when tackling complex real-world tasks where individual agents possess limited capabilities and restricted visibility of the environment \citep{lajoie2021towards, yu2022surprising, liu2021coach}.
In this work, we consider MARL scenarios wherein the task at hand is of a cooperative nature and agents are situated in a partially observable environment, but each is endowed with different observation power.
Each agent is underpinned by a deep feed-forward network, augmented with access to a communication channel conveying continuous vectors.
Considering bandwidth-related constraints, particularly in instances involving wireless communication channels, a limited subset of agents is permitted to exchange messages during each time step to ensure reliable message transfer \citep{kim2019learning}.
This necessitates meticulous consideration by agents in selecting both the information they convey and the recipient agent.

\begin{figure}
    \centering
    \includegraphics[width=1.0\linewidth]{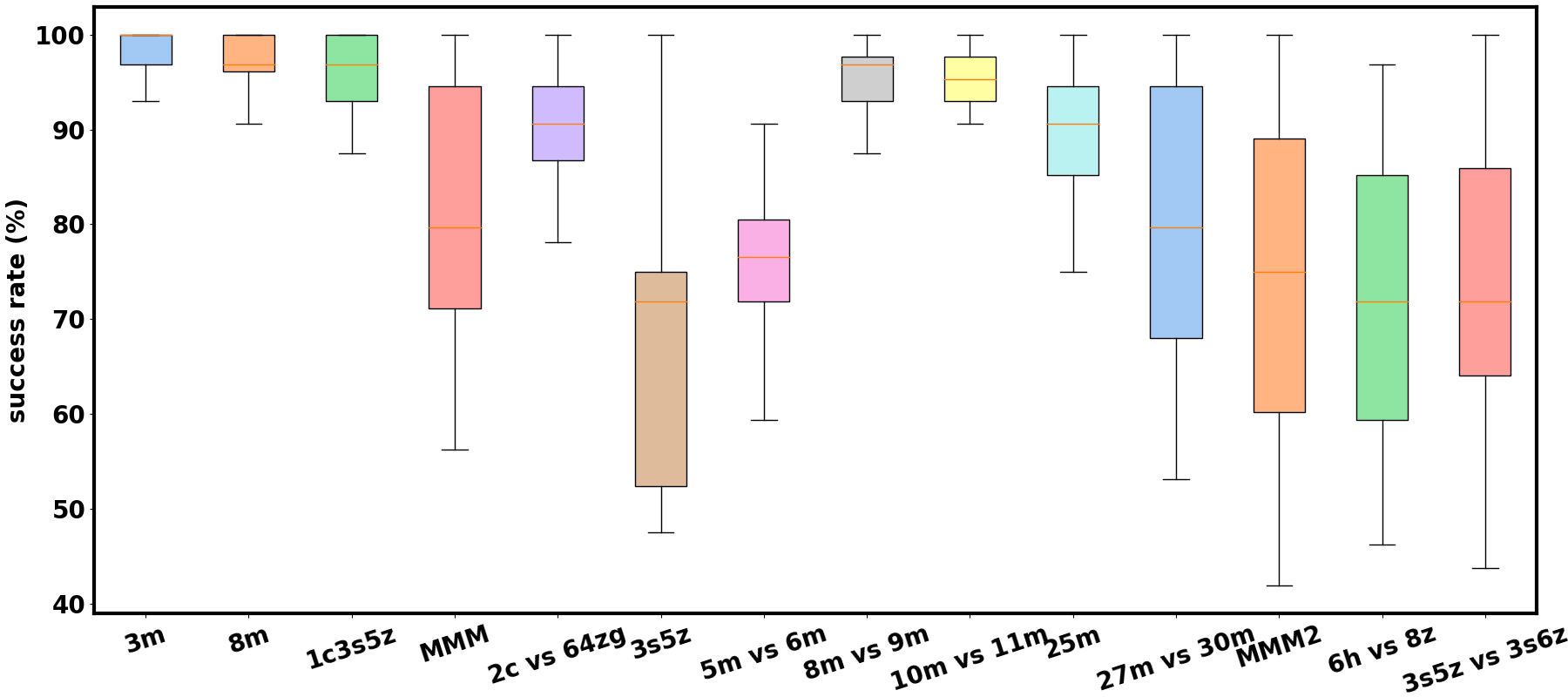}
     \vspace{-0.5cm}
    \caption{
    The performance of pre-defined communication architectures across various StarCraftII combat games, each with 10 different seeds. 
    The notable variance observed underscores the importance of searching for the optimal communication architecture.}
    \label{fig:variance}
    \vspace{-0.2cm}
\end{figure}

To facilitate coordinated message exchange, we adopt the centralized training and distributed execution paradigm, as popularized in recent works such as \citet{foerster2018counterfactual, kuba2021trust, mappo}, which allows agents access to global information and knowledge of opponents' actions during the training phase.
There are several approaches for learning communication in MARL including CommNet \citep{sukhbaatar2016learning}, TarMAC \citep{das2019tarmac}, and ToM2C \citep{wang2021tom2c}.
However, methods relying on information sharing among all agents or relying on manually pre-defined communication architectures can be problematic.
When dealing with a large number of agents, distinguishing valuable information for cooperative decision-making from globally shared data becomes problematic.
In such cases, communication may provide limited benefit and could potentially hinder cooperative learning \citep{jiang2018learning}. 
Furthermore, in real-world applications, full-scale communication between all agents can be costly, demanding high bandwidth, incurring delays, and imposing significant computational complexity.
Manual pre-defined architectures exhibit substantial variance, as evident in Figure \ref{fig:variance}, which underscores the necessity for meticulous architectural design to achieve optimal communication, as randomly designed architectures may inadvertently hinder cooperation and result in poor overall performance.
Dynamic adjustments to the communication graph during inference have garnered significant attention in recent research \citep{jiang2018learning, kim2019learning, wang2021tom2c}.
However, this approach assumes all agents always need to communicate with one of the other agents, necessitating complex scheduling algorithms, which results in the waste of bandwidth consumption and falls outside the scope of this article.

To address these challenges, we present a novel approach, named CommFormer, designed to facilitate effective and efficient communication among agents in large-scale MARL within partially observable distributed environments.
We conceptualize the communication structure among agents as a learnable graph and formulate this problem as the task of determining the communication graph while enabling the architecture parameters to update normally, thus necessitating a bi-level optimization process.
In contrast to conventional methods that involve searching through a discrete set of candidate communication architectures, we relax the search space into a continuous domain, enabling architecture optimization via gradient descent in an end-to-end manner.
Diverging from previous approaches that often employ arithmetic or weighted means of internal states before message transmission \citep{peng2017multiagent, wang2021tom2c}, which may compromise communication effectiveness, our method directly transmits each agent's local observations and actions to specific agents based on the learned communication architecture.
Subsequently, each agent employs an attention unit to dynamically allocate credit to received messages from the graph modeling perspective, which enjoys a monotonic performance improvement guarantee \citep{wen2022multi}.
Extensive experiments conducted in a variety of cooperative tasks substantiate the robustness of our model across diverse cooperative scenarios. 
CommFormer consistently outperforms strong baselines and achieves comparable performance to methods allowing information sharing among all agents, demonstrating its effectiveness regardless of variations in the number of agents.

Our contributions can be summarized as follows:

\begin{itemize}[leftmargin=*]
    \item We conceptualize the communication structure as a graph and introduce an innovative algorithm for learning it through bi-level optimization, which efficiently enables the simultaneous optimization of the communication graph and architectural parameters.
    \item We propose the adoption of the attention unit within the framework of graph modeling to dynamically allocate credit to received messages, thereby enjoying a monotonic performance improvement guarantee while also improving communication efficiency. 
    \item Through extensive experiments on a variety of cooperative tasks, CommFormer consistently outperforms robust baseline methods and achieves performance levels comparable to approaches that permit unrestricted information sharing among all agents.
\end{itemize}

\begin{figure}
    \centering
    \includegraphics[width=1.0\linewidth]{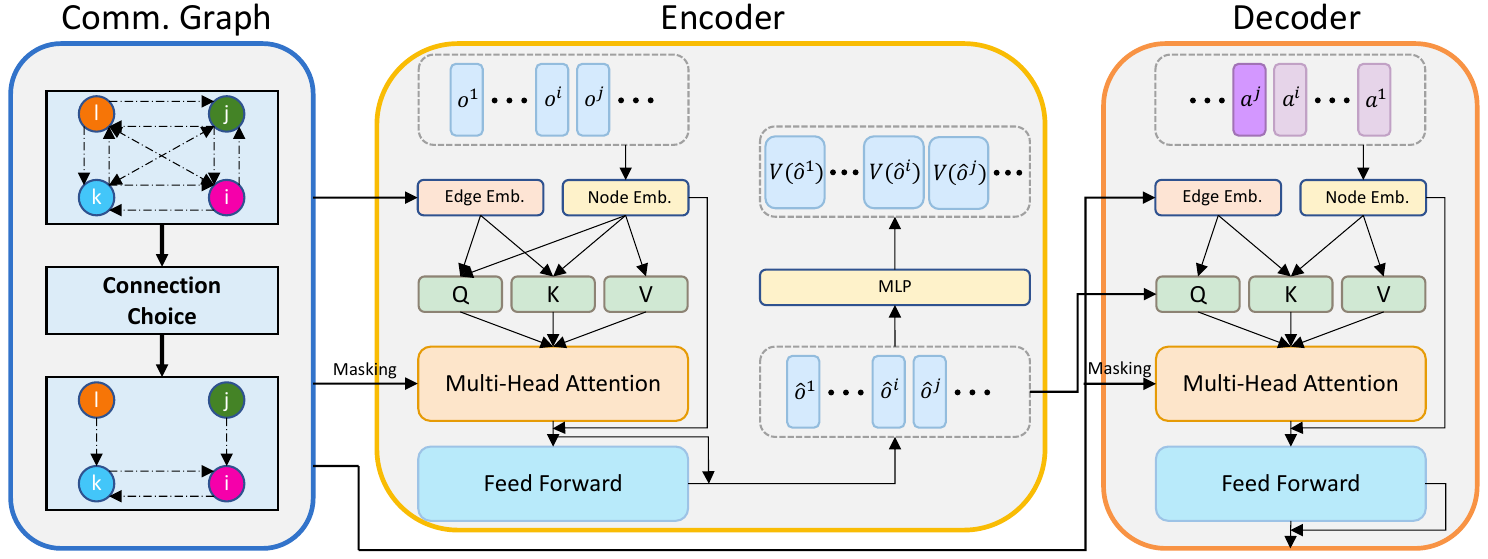}
    \caption{
    The overview of our proposed CommFormer. 
    CommFormer initiates by establishing the communication graph, which subsequently serves as both the masking and edge embeddings in the encoder and decoder to ensure that agents can exclusively access messages from communicated agents.
    Subsequently, the encoder and decoder modules come into play, processing a sequence of agents' observations and transforming them into a sequence of optimal actions.
    }
    \label{fig:overview}
    \vspace{-0.2cm}
\end{figure}
\section{Related Work}

\textbf{Multi-agent Cooperation}.
As a natural extension of single-agent RL, MARL has garnered considerable attention for addressing complex problems within the framework of Markov Games \citep{yang2020overview}.
Numerous MARL methodologies have been developed to tackle cooperative tasks in an online setting, where all participating agents collaborate toward a shared reward objective.
To address the challenge of non-stationarity in MARL, algorithms typically operate within two overarching frameworks: centralized and decentralized learning.
Centralized methods \citep{claus1998dynamics} involve the direct learning of a single policy responsible for generating joint actions for all agents.
On the other hand, decentralized learning \citep{littman1994markov} entails each agent independently optimizing its own reward function. 
While these methods can handle general-sum games, they may encounter instability issues, even in relatively simple matrix games \citep{foerster2017learning}.
Centralized training and decentralized execution (CTDE) algorithms represent a middle ground between these two frameworks. 
One category of CTDE algorithms is value-decomposition (VD) methods, wherein the joint Q-function is formulated as a function dependent on the individual agents' local Q-functions \citep{rashid2020monotonic, son2019qtran, sunehag2017value}.
The others \citep{lowe2017multi, foerster2018counterfactual} employ actor-critic architectures and learn a centralized critic that takes global information into account.
In this work, we introduce an innovative approach operating under the CTDE paradigm, with limited communication capabilities, which achieves comparable or even superior performance when compared to these established baselines.

\textbf{Communication Learning}.
Learning to facilitate communication is a viable approach to enhance multi-agent cooperation.
DIAL \citep{foerster2016learning} is the pioneer in proposing learnable communication through back-propagation with deep Q-networks. 
In this method, each agent generates a message at each timestep, which serves as input for other agents in the subsequent timestep. 
Building upon this work, a variety of approaches have emerged in the field of multi-agent communication.
Some methods adopt pre-defined communication architectures, e.g. CommNet \citep{sukhbaatar2016learning}, BiCNet \citep{peng2017multiagent}, and GA-Comm \citep{liu2020multi}. 
These techniques establish fixed communication structures to facilitate information exchange among agents, often utilizing GNN models \citep{niu2021multi, bettini2023heterogeneous}.
In contrast, other approaches such as ATOC \citep{jiang2018learning}, TarMAC \citep{das2019tarmac}, and ToM2C \citep{wang2021tom2c} explore dynamic adaptation of communication structures during inference, enabling agents to selectively transmit or receive information.
In our research, we align with the former approach, establishing a fixed communication architecture pre-inference. Each agent transmits its local observation and action as a message to a shared channel. Our novel CommFormer approach extends this concept by learning an optimal communication architecture through back-propagation. 
In contrast to CDC \citep{pesce2023learning}, which dynamically alters the communication graph through a diffusion process perspective, and TWG-Q \citep{liu2022temporal}, which emphasizes temporal weight learning and the application of weighted GCN, CommFormer adopts a different approach. It focuses on learning a static graph, aimed at optimizing communication efficiency prior to the inference phase, setting it apart from the traditional methodologies employed by the aforementioned approaches.

\section{Method}

The goal of our proposed method is to address the multi-agent collaborative communication problem, which enables agents to operate cohesively as a collective entity rather than disparate individuals.
In this paper, we are specifically interested in \textit{learning to construct the communication graph} and \textit{learning how to cooperate with received messages} in a bandwidth-limited way.

\subsection{Problem Formulation}
The MARL problems can be modeled by Markov games $\langle \mathcal{N}, \bm{\mathcal{O}}, \bm{\mathcal{A}}, R, P, \gamma \rangle$ \citep{littman1994markov}.
The set of agents is denoted as $\mathcal{N} = \{1, \dots, N \}$. 
The product of the local observation spaces of the agents forms the joint observation space, denoted as $\bm{\mathcal{O}}=\prod_{i=1}^{n}\mathcal{O}^i$. 
Similarly, the product of the agents' action spaces constitutes the joint action space, represented as $\bm{\mathcal{A}}=\prod_{i=1}^{n}\mathcal{A}^i$. 
The joint reward function, $R:\bm{\mathcal{O}}\times \bm{\mathcal{A}} \rightarrow [-R_{\max}, R_{\max}]$, maps the joint observation and action spaces to the reward range $[-R_{\max}, R_{\max}]$. The transition probability function, $P:\bm{\mathcal{O}}\times \bm{\mathcal{A}} \times \bm{\mathcal{O}} \rightarrow \mathbb{R}$, defines the probability distribution of transitioning from one joint observation and action to another. 
Lastly, the discount factor, denoted as $\gamma \in [0, 1)$, plays a crucial role in discounting future rewards.

At time step $t\in\mathbb{N}$, an agent $i\in\mathcal{N}$ receives an observation denoted as $\ro_{t}^i\in\mathcal{O}^i$. The collection of these individual observations $\vo=(o^1, \dots, o^n)$ forms the "joint" observation. Agent $i$ then selects an action $\ra^{i}_t$ based on its policy $\pi^i$. It's worth noting that $\pi^i$ represents the policy of the $i^{\text{th}}$ agent, which is a component of the agents' joint policy denoted as $\pi$.
Apart from its own local observation $\ro_{t}^i$, each agent possesses the capability to receive observations $\ro_t^j$ from other agents, along with their actions (auto-regressively) $\ra^{j}_t$ through a communication channel. 
At the end of each time step, the entire team collectively receives a joint reward denoted as $R(\rvo_{t}, \rva_t)$ and observes $\rvo_{t+1}$, following a probability distribution $P(\cdot|\rvo_{t}, \rva_t)$.
Over an infinite sequence of such steps, the agents accumulate a discounted cumulative return denoted as $R^{\gamma} \triangleq \sum_{t=0}^{\infty}\gamma^t R(\rvo_{t}, \rva_t)$.

In practical scenarios where agents have the capability to communicate with each other over a shared medium, two critical constraints are imposed: bandwidth and contention for medium access \citep{kim2019learning}.
The bandwidth constraint implies that there is a limited capacity for transmitting bits per unit time, and the contention constraint necessitates the avoidance of collisions among multiple transmissions, which is a natural aspect of signal broadcasting in wireless communication. 
Consequently, each agent can only transmit their message to a restricted number of other agents during each time step to ensure reliable message transfer.
In this paper, we conceptualize the communication architecture as a directed graph, denoted as $\mathcal{G} = \langle \mathcal{V}, \mathcal{E} \rangle$, where each node $v_i \in \mathcal{V}$ represents an agent, and an edge $e_{i \rightarrow j} \in \mathcal{E}$ signifies message passing from agent $v_i$ to agent $v_j$.
The restriction on communication can be mathematically expressed as the sparsity $\mathcal{S}$ of the adjacency matrix of the edge connections $\alpha$. 
This sparsity parameter, $\mathcal{S}$, controls the allowed number of connected edges, which is given by $\mathcal{S} \times N^2$, where $N$ is the number of agents.

\subsection{Architecture}
The overall architecture of our proposed CommFormer is illustrated in Figure \ref{fig:overview}.

\textbf{Communication Graph.} To design a communication-efficient MARL paradigm, we introduce the Communication Transformer or CommFormer, which adopts a graph modeling paradigm, inspired by developments in sequence modeling \citep{hu2023graph, wen2022multi}
We apply the Transformer architecture which facilitates the mapping between the input, consisting of agents' observation sequences $(o^{1}, \dots, o^{n})$, and the output, which comprises agents' action sequences $(a^{1}, \dots, a^{n})$.
Considering communication constraints, each agent has a limited capacity to communicate with a subset of other agents, represented by the sparsity $\mathcal{S}$ of the adjacency matrix of the edge connections.
To identify the optimal communication graph, we treat multiple agents as nodes in a graph and introduce a learnable adjacency matrix, represented by the parameter matrix $\alpha \in \mathbb{R}^{N \times N}$, which are optimized during training in an end-to-end manner.

\textbf{Encoder.} 
The encoder, whose parameters are denoted by $\phi$, takes a sequence of observations $(o^{1}, \dots, o^{n})$ as input and passes them through several computational blocks.
Each such block consists of a \textsl{relation-enhanced} mechanism \citep{hu2023graph, cai2020graph} and a \textsl{multi-layer perceptron} (MLP), as well as \textsl{residual connections} to prevent gradient vanishing and network degradation with the increase of depth. 
In the vanilla multi-head attention, the attention score between the element $o^i$ and $o^j$ can be formulated as the dot-product between their query vector and key vector:
\begin{equation}
    s_{ij} = f(o^i, o^j) = o^i W_q^T W_k o^j.
\end{equation}
$s_{ij}$ can be regarded as implicit information associated with the edge $e_{j \rightarrow i}$, where agent $o^i$ queries the information sent from agent $o^j$.
To identify the most influential edge contributing to the final performance, we augment the implicit attention score with explicit edge information:
\begin{equation}
\label{eq:attn_score}
    \begin{split}
        s_{ij} &= g(o^i, o^j, r_{i \rightarrow j}, r_{j \rightarrow i}) \\
            &= (o^i + r_{i \rightarrow j})W_q^TW_k (o^j + r_{j \rightarrow i} ),
    \end{split}
\end{equation}
where $r_{* \rightarrow *}$ is obtained from an embedding layer that takes the adjacency matrix $\alpha$ as input.
We also apply a mask to the attention scores using the adjacency matrix $\alpha$ to ensure that only information from connected agents is accessible:
\begin{equation}
\label{eq:mask}
    s_{ij} = 
    \begin{cases}
        s_{ij}, & e_{j\rightarrow i} = 1, \\
        -\infty, & e_{j\rightarrow i} = 0. 
    \end{cases}
\end{equation}
We represent the encoded observations as $(\hat{\vo}^{1}, \dots, \hat{\vo}^{n})$, which capture not only the individual agent information but also the higher-level inter-dependencies between agents through communication.
To facilitate the learning of expressive representations, during the training phase, we treat the encoder as the critic and introduce an additional projection to estimate the value functions:
\begin{equation}
    L_{\text{Encoder}}(\phi) = \frac{1}{Tn}\sum_{m=1}^{n}\sum_{t=0}^{T-1}\Big[ R(\rvo_t, \rva_t) + \gamma V_{\bar{\phi}}(\hat{\rvo}^{{m}}_{t+1}) - V_{\phi}(\hat{\rvo}^{{m}}_t)\Big]^2,
    \label{eq:encoder-loss}
\end{equation}
where $\bar{\phi}$ is the target network's parameter, which is a separate neural network that is a copy of the main value function. 
The update mechanism for $\bar{\phi}$ is executed either through an exponential moving average or via periodic updates in a "hard" manner \citep{mnih2015human}.

\textbf{Decoder.} 
The decoder, characterized by its parameters $\theta$, processes the embedded joint action $\va^{{0:m-1}}, m={1, \dots n}$ through a series of decoding blocks.
The decoding block also incorporates a \textsl{relation-enhanced} mechanism for calculating attention between encoded actions and observation representations, along with an MLP and \textsl{residual connections}.
In addition to the adjacency matrix mask, we apply a constraint that limits attention computation to occur only between agent $i$ and its preceding agents $j$ where $j < i$. 
This constraint maintains the sequential update scheme, ensuring that the decoder produces the action sequence in an auto-regressive manner: $\pi^{m}_{\theta}(\ra^{m}|\hat{\rvo}^{{1:n}}, \rva^{{1:m-1}})$, which guarantees monotonic performance improvement during training \citep{wen2022multi}.
We apply the PPO algorithm \citep{schulman2017proximal} to train the decoder agent:
\begin{equation}
\label{eq:decoder-loss}
    \begin{split}
        L_{\text{Decoder}}(\theta) &= -\frac{1}{Tn}\sum_{m=1}^{n}\sum_{t=0}^{T-1}\min\Big( \rr^{m}_{t}(\theta)\hat{A}_t, \text{clip}(\rr^{m}_{t}(\theta), 1\pm \epsilon)\hat{A}_t \Big), \\
        \rr^{m}_{t}(\theta) &= \frac{\pi^{m}_{\theta}(\ra^{m}_t|\hat{\rvo}^{1:n}_t, \hat{\rva}^{1:m-1}_t)}{\pi^{m}_{\theta_{\text{old}}}(\ra^{m}_t|\hat{\rvo}^{1:n}_t, \hat{\rva}^{1:m-1}_t)}, 
    \end{split}
\end{equation}
where $\hat{A}_t$ is an estimate of the joint advantage function, which can be formulated as $\hat{V}_t = \frac{1}{n}\sum_{m=1}^{n}V(\hat{\ro}^{m}_t)$ \citep{schulman2015high}.

\subsection{Training and Execution}

We employ the CTDE paradigm: during centralized training, there are no restrictions on communication between agents.
However, once the learned policies are executed in a decentralized manner, agents can only communicate through a constrained bandwidth channel.

\subsubsection{Centralized Training}

During the training stage, we need to determine the communication matrix $\alpha$ while allowing the architecture parameters $\phi$ and $\theta$ to update normally.
This implies a bi-level optimization problem \citep{anandalingam1992hierarchical, colson2007overview} with $\alpha$ as the upper level variable and $\phi$ and $\theta$ as the lower-level variable:
\begin{align}
    \min_{\alpha}~~~ & \mathcal{L}_{val} (\phi^*(\alpha), \theta^*(\alpha), \alpha), \\
    \text{s.t.}~~~ & \phi^*(\alpha), \theta^*(\alpha) = \argmin_{\phi, \theta} \mathcal{L}_{train}(\phi, \theta, \alpha) \label{eq_inner}, \\
    & |\alpha| \leq \mathcal{S} \times N^2 \label{eq_sparsity},
\end{align}
where $\mathcal{L} = L_{\text{Encoder}}(\phi) + L_{\text{Decoder}}(\theta)$ with different online rollouts for training $L_{train}$ and validation $L_{val}$, and $|\alpha|$ denotes the number of connected edges.
Evaluating the architecture gradient exactly can be prohibitive due to the expensive inner optimization, and each value in $\alpha$ is represented by a discrete value in $\{0, 1\}$.
We propose a simple approximation scheme that alternately updates the following formula and relaxes $\alpha$ as a continuous matrix to enable differentiable updating:
\begin{equation}
\label{eq:weight_loss}
        \phi = \phi - \xi \nabla_{\phi} \mathcal{L}_{train}(\phi, \theta, \alpha), ~ 
    \theta = \theta - \xi \nabla_{\theta} \mathcal{L}_{train} (\phi, \theta, \alpha),
\end{equation}
and 
\begin{equation}
\label{eq:structure_loss}
    \alpha = \alpha - \eta \nabla_{\alpha} \mathcal{L}_{val} (\phi, \theta, \alpha),
\end{equation}
where $\phi, \theta$ denote the current weights maintained by the algorithm, and $\xi, \eta$ are the learning rate for a step of inner and outer optimization.
The idea is to approximate $\phi^*(\alpha), \theta^*(\alpha)$ by adapting $\phi$ and $\theta$ using only a single training step, without fully solving the inner optimization (Equation \ref{eq_inner}) by training until convergence.

To update the discrete adjacency matrix $\alpha$, we utilize the Gumbel-Max trick \citep{jang2016categorical, maddison2016concrete} to sample the binary adjacency matrix, which facilitates the continuous representation of $\alpha$ and enables the normal back-propagation of gradients during training.
To satisfy constraint \ref{eq_sparsity}, we extend the original one-hot Gumbel-Max trick to k-hot, enabling each agent to send messages to a fixed number of $k = \mathcal{S} \times N$ agents:
\begin{equation}
\label{eq:sample_gumbel}
    e_i = \verb|k_hot| \big( \text{k-}\argmax \left[\text{Softmax}(\alpha_{ij} + g_j), ~ \text{for} ~ j=1,\dots,n \right] \big), 
\end{equation}
where $g_j$ is sampled from Gumbel(0,1), and $e_i \in \mathbb{N}^N$ represents the edges connected to agent $i$.
%

\subsubsection{Distributed Execution}

During execution, each agent $i$ has access to its local observations and actions, as well as additional information transmitted by other agents through communication.
The adjacency matrix is derived from the parameters $\alpha$ without any randomness as follows:
\begin{equation}
\label{eq:sample_exact}
    e_i = \verb|k_hot| \big( \text{k-}\argmax \left[\alpha_{ij}, ~ \text{for} ~ j=1,\dots,n \right] \big).
\end{equation}
Note that each action is generated auto-regressively, in the sense that $a^m$ will be inserted back into the decoder again to generate $a^{m+1}$ (starting with $a^0$ and ending with $a^{n-1}$).
Through the use of limited communication, each agent is still able to effectively select actions when compared to fully connected agents, which leads to significant reductions in communication costs and overhead.
The overall pseudocode is presented in Algorithm \ref{alg:commformer}.

\vspace{-0.2cm}
\begin{algorithm}[!htbp]
    \caption{CommFormer}
    \begin{algorithmic}[1]
    
    \STATE \textbf{Input:} Batch size $B$, number of agents $N$, episodes $K$, steps per episode $T$, sparsity $\mathcal{S}$.\\
    \STATE \textbf{Initialize:} Encoder $\{\phi\}$, Decoder $\{\theta\}$, Replay buffer $\mathcal{B}$, Adjacency matrix $\alpha \in \mathbb{R}^{n \times n}$. \\
    
    \FOR{$k = 0,1, \dots, K-1$}
        \FOR{$t = 0,1, \dots, T-1$}
            \STATE Collect a sequence of observations $o^{{1}}_t, \dots, o^{{n}}_t$ from environments.\\
            \STATE {\color{gray}\te{// inference with CommFormer}} \\
            \STATE Generate the matrix $e \in \{ 0, 1\}^{n \times n}$ according to the $\alpha$ with Equation \ref{eq:sample_exact}.
            \STATE Generate representation sequence $\hat{o}^{{1}}_t, \dots, \hat{o}^{{n}}_t$ via Encoder $\phi$ with attention score (Equation \ref{eq:attn_score}) and mask (Equation \ref{eq:mask}), similar to the Decoder. \\
            \FOR{$m = 0, 1, \dots, n-1$}
                \STATE Input $\hat{o}^{{1}}_t, \dots, \hat{o}^{{n}}_t$ and $a^{0}_t, \dots, a^{m}_t$ to the Decoder $\theta$ and infer $a^{{m+1}}_t$ with the auto-regressive manner. \\
            \ENDFOR
            \STATE Execute joint actions $a^{0}_t, \dots, a^{n}_t$ in environments and collect the reward $R(\vo_t, \va_t)$.\\
            \STATE Insert $(\vo_{t}, \va_{t},R (\vo_t, \va_t))$ in to $\mathcal{B}$.\\
        \ENDFOR\\
        \STATE {\color{gray}\te{// train the CommFormer}} \\
        \STATE Sample a random minibatch of $B$ steps from $\mathcal{B}$.
        \STATE Generate the matrix $e \in \{ 0, 1\}^{n \times n}$ according to the $\alpha$ with Equation \ref{eq:sample_gumbel}.
        \STATE Generate $V_\phi(\hat{o})$ with the output layer of the Encoder $\phi$ and compute the joint advantage function $\hat{A}$ based on $V_\phi(\hat{o})$ with GAE.\\
        \STATE Input $\hat{o}^{{1}}, \dots, \hat{o}^{{n}}$ and $a^{0}, \dots, a^{{n-1}}$, generate $\pi^{1}_{\theta}, \dots, \pi^{n}_{\theta}$ at once with the Decoder $\theta$.\\
        \STATE Calculate the training loss $L = L_{\text{Encoder}}(\phi) + L_{\text{Decoder}}(\theta)$ with Equation 
        \ref{eq:encoder-loss} and Equation (\ref{eq:decoder-loss}).\\
        \STATE Iteratively update the $\phi, \theta$ and $\alpha$ with Equation \ref{eq:weight_loss} and Equation \ref{eq:structure_loss}.
    \ENDFOR
    \end{algorithmic}
    \label{alg:commformer}
\end{algorithm}

\section{Experiment}

To evaluate the properties and performance of our proposed CommFormer\footnote{Our code is available at: \url{https://github.com/charleshsc/CommFormer}}, we conduct a series of experiments using four environments, including Predator-Prey (PP) \citep{singh2018learning}, Predator-Capture-Prey (PCP) \citep{seraj2022learning}, StarCraftII Multi-Agent Challenge (SMAC) \citep{samvelyan2019starcraft}, and Google Research Football(GRF) \citep{kurach2020google}.
A comprehensive description of each environment can be found in the Appendix \ref{sec:detail_env}.
It is worth noting that in certain domains, our objective extends beyond maximizing the average success rate or cumulative rewards. We also aim to minimize the average number of steps required to complete an episode, emphasizing the ability to achieve goals in the shortest possible time.

\subsection{Baselines}
We compare CommFormer with strong CTDE baselines that do not involve communication, e.g. HAPPO \citep{mappo}, MAPPO \citep{mappo} and QMIX \citep{rashid2020monotonic}, as well as popular communication methods, e.g. MAGIC \citep{niu2021multi}, TarMAC \citep{das2019tarmac}, and QGNN \citep{kortvelesy2022qgnn} to highlight its effectiveness. 
Details for each method are provided in Appendix \ref{sec:baselines}.
During experiments, the implementations of baseline methods are consistent with their official repositories, all hyper-parameters left unchanged at the origin best-performing status. 
We also include the \textbf{fully connected CommFormer} (FC) configuration, where there are no limitations on communication bandwidth.
In this configuration, each agent can communicate with all other agents, implying that the sparsity parameter $\mathcal{S}$ is set to 1.
FC serves as the upper bound of our methods and demonstrates strong performance on cooperative MARL tasks.

\subsection{Main Results}

\begin{table}[t!]
\caption{
Performance evaluations of different metrics and standard deviation on the selected benchmark, where UPDeT's official codebase supports several Marine-based tasks only.
Note that the sparsity parameter $\mathcal{S}$ in CommFormer is consistently set to 0.4 for all tasks.
}
 \renewcommand{\arraystretch}{1.0}
  \centering
    \resizebox{0.97\textwidth}{!}{
    \begin{tabular}{cc|ccccccc|c}
    \toprule
    Task & Difficulty & CommFormer(0.4) &  MAT & MAPPO & HAPPO & QMIX & UPDeT &FC & Steps\\
    \midrule
    \tiny{3m} & Easy & \textbf{100.0}\tiny{(0.0)} &  \textbf{100.0}\tiny{(0.0)} & \textbf{100.0}\tiny{(0.4)} & \textbf{100.0}\tiny{(1.2)} & 96.9\tiny{1.3} & \textbf{100.0}\tiny{(5.2)} & \textbf{100.0}\tiny{(0.0)} & 5e5\\
    \tiny{8m} & Easy &  \textbf{100.0}\tiny{(0.0)} & \textbf{100.0}\tiny{(0.0)} & 96.8\tiny{(2.9)} & 97.5\tiny{(1.1)} & 97.7\tiny{1.9} & 96.3\tiny{(9.7)}& \textbf{100.0}\tiny{(0.0)} & 1e6\\
    \tiny{1c3s5z} & Easy &  \textbf{100.0}\tiny{(0.0)} & \textbf{100.0}\tiny{(0.0)} & \textbf{100.0}\tiny{(2.2)} & 97.5\tiny{(1.8)} & 96.9\tiny{(1.5)} & / & \textbf{100.0}\tiny{(0.0)} & 2e6\\
    \tiny{MMM} & Easy & \textbf{100.0}\tiny{(0.0)}  & 83.3\tiny{(4.8)} & 95.6\tiny{(4.5)} & 81.2\tiny{(22.9)} & 91.2\tiny{(3.2)} & /& \textbf{100.0}\tiny{(0.0)} & 2e6\\
    \tiny{2c vs 64zg} & Hard & \textbf{100.0}\tiny{(0.0)}  &  \textbf{100.0}\tiny{(3.1)} & \textbf{100.0}\tiny{(2.7)} & 90.0\tiny{(4.8)} & 90.3\tiny{(4.0)} & / & \textbf{100.0}\tiny{(3.1)} & 5e6\\
    \tiny{3s5z} & Hard & \textbf{100.0}\tiny{(0.0)} & 74.0\tiny{(6.4)} & 72.5\tiny{(26.5)} & 90.0\tiny{(3.5)} & 84.3\tiny{(5.4)} & / & \textbf{100.0}\tiny{(3.1)} & 3e6\\
    \tiny{5m vs 6m} & Hard & 89.6\tiny{(1.5)} & 81.3\tiny{(5.1)} & 88.2\tiny{(6.2)} & 73.8\tiny{(4.4)} & 75.8\tiny{(3.7)} & 90.6\tiny{(6.1)}& \textbf{93.8}\tiny{(4.4)} & 1e7\\
    \tiny{8m vs 9m} & Hard & \textbf{100.0}\tiny{(0.0)} &  96.9\tiny{(0.0)} & 93.8\tiny{(3.5)} & 86.2\tiny{(4.4)} & 92.6\tiny{(4.0)} & /& \textbf{100.0}\tiny{(3.1)} & 5e6\\
    \tiny{10m vs 11m} & Hard & \textbf{100.0}\tiny{(1.4)} & \textbf{100.0}\tiny{(3.1)} & 96.3\tiny{(5.8)} & 77.5\tiny{(9.7)} & 95.8\tiny{(6.1)} & / & \textbf{100.0}\tiny{(0.0)} & 5e6\\
    \tiny{25m} & Hard & \textbf{100.0}\tiny{(0.0)} &  0.0\tiny{(0.1)} & \textbf{100.0}\tiny{(2.7)} & 0.6\tiny{(0.8)} & 90.2\tiny{(9.8)} & 2.8\tiny{(3.1)}& \textbf{100.0}\tiny{(0.0)} & 2e6\\
    \tiny{27m vs 30m} & Hard+ & 96.9\tiny{(3.1)} & 80.2\tiny{(4.8)} & 93.1\tiny{(3.2)} & 0.0\tiny{(0.0)} & 39.2\tiny{(8.8)} & /& \textbf{100.0}\tiny{(0.0)} & 1e7\\
    \tiny{MMM2} & Hard+ & \textbf{100.0}\tiny{(3.1)} & 96.9\tiny{(0.0)} & 81.8\tiny{(10.1)} & 0.3\tiny{(0.4)} & 88.3\tiny{(2.4)} & / & \textbf{100.0}\tiny{(0.0)} & 1e7\\
    \tiny{6h vs 8z} & Hard+ & 96.9\tiny{(3.1)}  & 93.8\tiny{(4.4)} & 88.4\tiny{(5.7)} & 0.0\tiny{(0.0)} & 9.7\tiny{(3.1)} & /& \textbf{100.0}\tiny{(0.0)} & 1e7\\
    \tiny{3s5z vs 3s6z} & Hard+ & 87.5\tiny{(3.1)} & 79.2\tiny{(9.0)} & 84.3\tiny{(19.4)} & 82.8\tiny{(21.2)} & 68.8\tiny{(21.2)} & /& \textbf{100.0}\tiny{(3.1)} & 2e7\\
    \bottomrule
    \toprule
    Task & Difficulty & CommFormer(0.4) & QGNN & SMS & TarMAC & NDQ & MAGIC & QMIX  & Steps\\
    \midrule
    \tiny{1o2r vs 4r} & Hard+ & \textbf{96.9}\tiny{(1.5)} & 93.8\tiny{(2.6)} & 76.4 & 39.1 & 77.1 & 22.3 & 51.1  & 2e7  \\
    \tiny{5z vs 1ul} & Hard+ & \textbf{100.0}\tiny{(1.4)} & 92.2\tiny{(1.6)} & 59.9 & 44.2 & 48.9 & 0.0 & 82.6 & 1e7 \\
    \tiny{1o10b vs 1r} & Hard+ & 96.9\tiny{(3.1)} & \textbf{98.0}\tiny{(2.9)} & 86.0 & 40.1 & 78.1 & 5.8 & 51.4 & 2e7 \\
    \bottomrule
    \toprule
    Task & Metric & CommFormer(0.4) & MAGIC & HetNet & CommNet & I3CNet & TarMAC & GA-Comm & Steps\\
    \midrule
     & Success Rate & \textbf{100.0}\tiny{(0.0)} & 98.2\tiny{(1.0)} & / & 59.2\tiny{(13.7)} & 70.0\tiny{(9.8)} & 73.5\tiny{(8.3)} & 88.8\tiny{(3.9)} & - \\
    
    \multirow{-2}{*}{GRF}& Steps Taken & \textbf{25.4}\tiny{(0.4)} & 34.3\tiny{(1.3)} & / & 39.3\tiny{(2.4)} & 40.4\tiny{(1.2)} & 41.5\tiny{(2.8)} & 39.1\tiny{(3.1)} & - \\ \midrule
     & Avg. Cumulative $\mathcal{R}$ & \textbf{-0.121}\tiny{(0.008)}  & -0.386\tiny{(0.024)}  & -0.232\tiny{(0.010)} & -0.336\tiny{(0.012)} & -0.342\tiny{(0.015)} & -0.563\tiny{(0.030)} & / & - \\
    \multirow{-2}{*}{PP}& Steps Taken & \textbf{4.99}\tiny{(0.31)} & 10.6\tiny{(0.50)} & 8.30\tiny{(0.25)} & 8.97\tiny{(0.25)} & 9.69\tiny{(0.26)} & 18.4\tiny{(0.46)} & / & - \\ \midrule
    & Avg. Cumulative $\mathcal{R}$ & \textbf{-0.197}\tiny{(0.019)} & -0.394\tiny{(0.017)} & -0.364\tiny{(0.017)} & -0.394\tiny{(0.019)} & -0.411\tiny{(0.019)} & -0.548\tiny{(0.031)} & / & - \\
    \multirow{-2}{*}{PCP}& Steps Taken & \textbf{7.61}\tiny{(0.66)} & 10.8\tiny{(0.45)} & 9.98\tiny{(0.36)} & 11.3\tiny{(0.34)} & 11.5\tiny{(0.37)} & 17.0\tiny{(0.80)} & / & - \\
    \bottomrule
    \end{tabular}
    }
\label{tab:smac}
\end{table}

According to the results presented in Table \ref{tab:smac} and Figure \ref{fig:variance}, our CommFormer with a sparsity parameter $\mathcal{S}=0.4$ significantly outperforms the state-of-the-art baselines. 
It consistently finds the optimal communication architecture across diverse cooperative scenarios, regardless of changes in the number of agents.
Take the task \textit{3s5z} as an example, where the algorithm needs to control different types of agents: stalkers and zealots.
This requires careful design of the communication architecture based on the capabilities of different units. 
Otherwise, it can even have a detrimental impact on performance, as indicated by the substantial variance displayed in Figure \ref{fig:variance}.
The outcome of \textit{3s5z} presented in Table \ref{tab:smac} consistently highlights CommFormer's ability to attain optimal performance with different random seeds, which underscores the robustness and efficiency of our proposed method.
Furthermore, in comparison to the FC method, CommFormer nearly matches its performance while retaining only 40\% of the edges. 
This indicates that with an appropriate communication architecture, many communication channels can be eliminated, thereby significantly reducing the hardware communication equipment requirements and expanding its applicability.
Finally, it's worth noting that all the results presented in Table \ref{tab:smac} are based on the same number of training steps, demonstrating the robustness and effectiveness of our bi-level optimization approach, which consistently converges to the optimal solution while maintaining sample efficiency. 
The hyper-parameters used in the study and additional detailed results can be found in the Appendix.

\subsection{Ablations}

We conduct several ablation studies, primarily focusing on the SMAC environments, to examine specific aspects of our CommFormer.
The parameter $\mathcal{S}$, which determines the sparsity of the communication graph, impacts the number of connected edges. 
Lower values of $\mathcal{S}$ imply reduced costs associated with communication but may also lead to performance degradation.
Additionally, we conduct ablation studies to investigate the essence of architecture searching, where we generate various pre-defined architectures using different random seeds, simulating manually pre-defined settings. 

\begin{figure}
    \centering
    \begin{subfigure}{0.32\linewidth}
        \centering
        \includegraphics[width=1.7in]{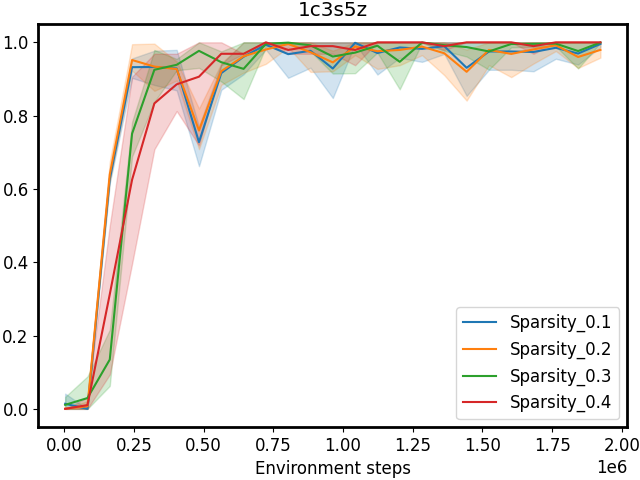} 
    \end{subfigure}%
    \begin{subfigure}{0.32\linewidth}
        \centering
        \includegraphics[width=1.7in]{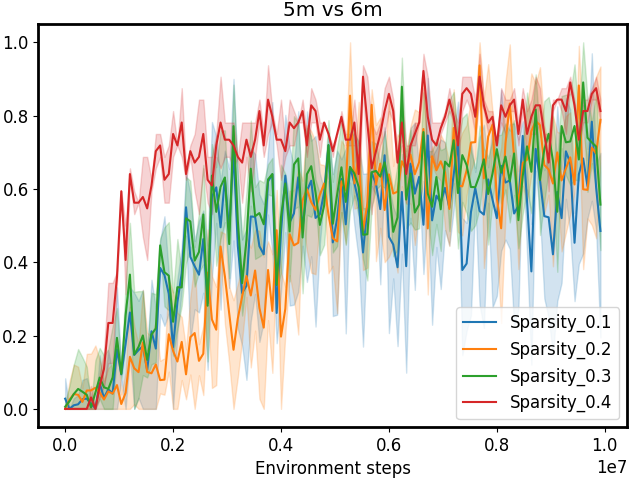} 
    \end{subfigure}%
    \begin{subfigure}{0.32\linewidth}
        \centering
        \includegraphics[width=1.7in]{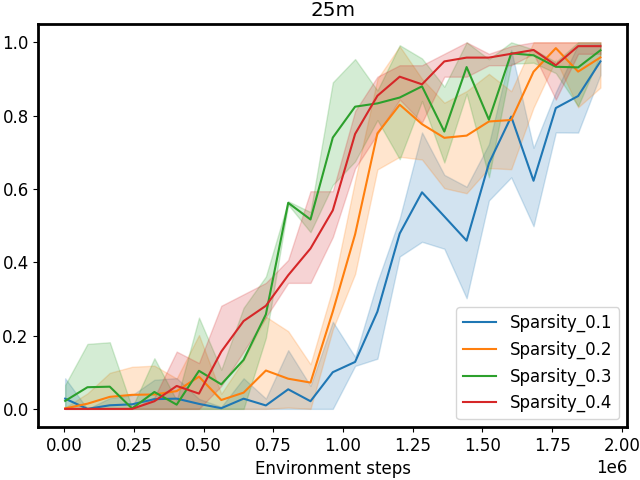} 
    \end{subfigure}%
    \caption{
    Performance comparison on SMAC tasks with different sparsity $\mathcal{S}$.
    Note that as the value of sparsity $\mathcal{S}$ gradually increases, the performance of CommFormer improves across various environments. This effect is particularly pronounced in environments with a large number of agents.
    }
    \label{fig:absparsity}
    \vspace{-0.3cm}
\end{figure}

\begin{figure}
    \centering
    \begin{subfigure}{0.32\linewidth}
        \centering
        \includegraphics[width=1.7in]{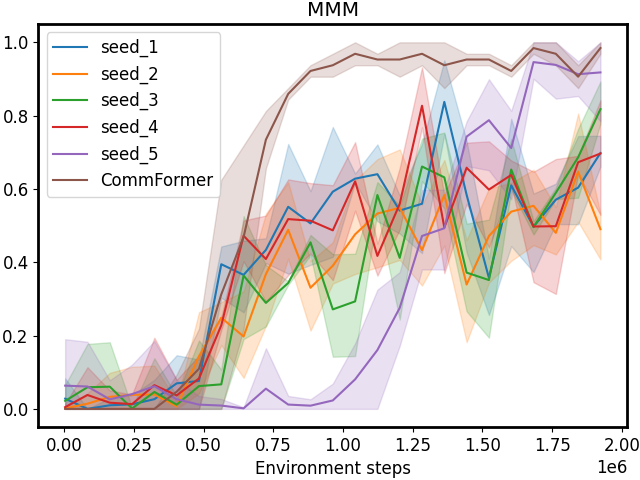} 
    \end{subfigure}%
    \begin{subfigure}{0.32\linewidth}
        \centering
        \includegraphics[width=1.7in]{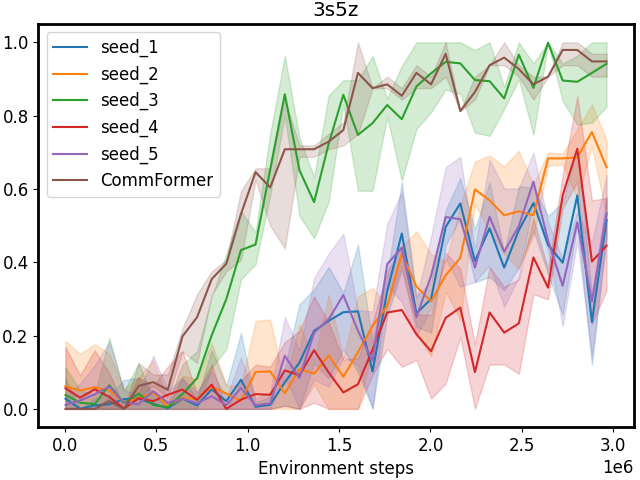} 
    \end{subfigure}%
    \begin{subfigure}{0.32\linewidth}
        \centering
        \includegraphics[width=1.7in]{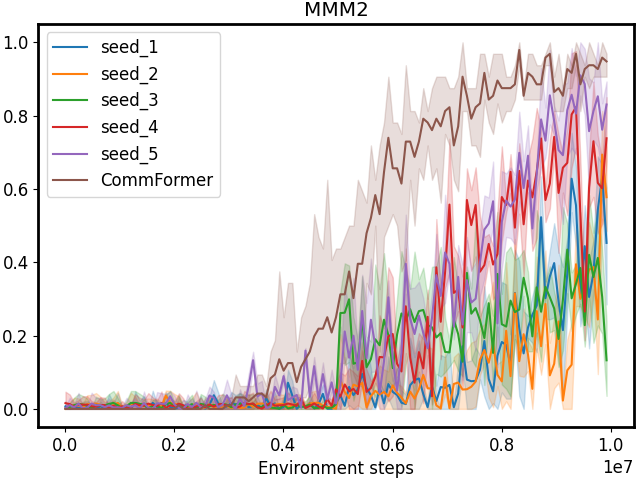} 
    \end{subfigure}%
    \caption{
    Performance comparison on SMAC tasks with different manually pre-defined communication architectures.
    CommFormer consistently achieves optimal performance, which underscores its capability to autonomously search for the optimal communication architecture, highlighting its adaptability across various scenarios and tasks.
    }
    \label{fig:absearch}
    \vspace{-0.3cm}
\end{figure}

\textbf{Sparsity.}
The parameter $\mathcal{S}$ introduced in our bi-level optimization controls the number of connected edges, ensuring that it does not exceed $\mathcal{S} \times N^2$, as specified in Equation \ref{eq_sparsity}. 
To simplify this constraint, we ensure that the total number of edges $|\alpha|$ equals $\mathcal{S} \times N^2$, with each agent communicating with a fixed number of $\mathcal{S} \times N$ agents.
Smaller values of $\mathcal{S}$ reduce the cost associated with communication but may also result in performance degradation. 
To investigate the impact of varying $\mathcal{S}$, we conduct a series of experiments, whose results are presented in Figure \ref{fig:absparsity}.
For simpler tasks, such as \textit{1c3s5z}, achieving a 100\% win rate is possible even when each agent can only communicate with one other agent. 
Nevertheless, As task complexity and the number of participating agents increase, a larger value of sparsity $\mathcal{S}$ becomes necessary to attain superior performance.

\textbf{Architecture Searching.} 
In light of the constraints imposed by limited bandwidth and contention for medium access, designing the communication architecture for each agent becomes a critical task.
To investigate the impact of different communication graph configurations, we conduct experiments using various random seeds, simulating different individuals' approaches to the problem. 
The results, as depicted in Figure \ref{fig:variance} and \ref{fig:absearch}, highlight that manually pre-defining the communication architecture often leads to significant performance variance, demanding expert knowledge for achieving better results.
In contrast, our proposed method leverages the continuous relaxation of the graph representation. This innovative approach allows for the simultaneous optimization of both the communication graph and architectural parameters in an end-to-end fashion, all while maintaining sample efficiency. 
This underscores the essentiality and effectiveness of our approach in tackling the challenges of multi-agent communication in constrained environments.

\section{Conclusion}

In this paper, we introduce a novel approach called CommFormer, which addresses the challenge of learning multi-agent communication from a graph modeling perspective.
Our approach treats the communication architecture among agents as a learnable graph and formulates this problem as the task of determining the communication graph while enabling the architecture parameters to update normally, thus necessitating a bi-level optimization process.
By leveraging continuous relaxation of graph representation and incorporating attention mechanisms within the graph modeling framework, CommFormer enables the concurrent optimization of the communication graph and architectural parameters in an end-to-end manner.
Extensive experiments conducted on a variety of cooperative tasks illustrate the significant performance advantage of our approach compared to other state-of-the-art baseline methods. 
In fact, CommFormer approaches the upper bound in scenarios where unrestricted information sharing among all agents is permitted. 
We believe that our work opens up new possibilities for the application of communication learning in the field of MARL, where effective communication plays a pivotal role in addressing various challenges.

\subsubsection*{Ethics Statements}
This paper does not raise any ethics concerns. This study does not involve any human subjects, practices to data set releases, potentially harmful insights, methodologies and applications, potential conflicts of interest and sponsorship, discrimination/bias/fairness concerns, privacy and security issues, legal compliance, and research integrity issues.

\subsubsection*{Acknowledgments}
This work is supported by the National Key R\&D Program of China (No. 2022ZD0160702), STCSM (No. 22511106101, No. 22511105700, No. 21DZ1100100), 111 plan (No. BP0719010) and National Natural Science Foundation of China (No. 62306178).

\bibliography{iclr2024_conference}

\begin{thebibliography}{50}
\providecommand{\natexlab}[1]{#1}
\providecommand{\url}[1]{\texttt{#1}}
\expandafter\ifx\csname urlstyle\endcsname\relax
  \providecommand{\doi}[1]{doi: #1}\else
  \providecommand{\doi}{doi: \begingroup \urlstyle{rm}\Url}\fi

\bibitem[Anandalingam \& Friesz(1992)Anandalingam and Friesz]{anandalingam1992hierarchical}
G~Anandalingam and Terry~L Friesz.
\newblock Hierarchical optimization: An introduction.
\newblock \emph{Annals of Operations Research}, 1992.

\bibitem[Bettini et~al.(2023)Bettini, Shankar, and Prorok]{bettini2023heterogeneous}
Matteo Bettini, Ajay Shankar, and Amanda Prorok.
\newblock Heterogeneous multi-robot reinforcement learning.
\newblock \emph{arXiv preprint arXiv:2301.07137}, 2023.

\bibitem[Cai \& Lam(2020)Cai and Lam]{cai2020graph}
Deng Cai and Wai Lam.
\newblock Graph transformer for graph-to-sequence learning.
\newblock In \emph{AAAI}, 2020.

\bibitem[Claus \& Boutilier(1998)Claus and Boutilier]{claus1998dynamics}
Caroline Claus and Craig Boutilier.
\newblock The dynamics of reinforcement learning in cooperative multiagent systems.
\newblock \emph{AAAI/IAAI}, 1998.

\bibitem[Colson et~al.(2007)Colson, Marcotte, and Savard]{colson2007overview}
Beno{\^\i}t Colson, Patrice Marcotte, and Gilles Savard.
\newblock An overview of bilevel optimization.
\newblock \emph{Annals of operations research}, 2007.

\bibitem[Das et~al.(2019)Das, Gervet, Romoff, Batra, Parikh, Rabbat, and Pineau]{das2019tarmac}
Abhishek Das, Th{\'e}ophile Gervet, Joshua Romoff, Dhruv Batra, Devi Parikh, Mike Rabbat, and Joelle Pineau.
\newblock Tarmac: Targeted multi-agent communication.
\newblock In \emph{ICML}, 2019.

\bibitem[Foerster et~al.(2016)Foerster, Assael, De~Freitas, and Whiteson]{foerster2016learning}
Jakob Foerster, Ioannis~Alexandros Assael, Nando De~Freitas, and Shimon Whiteson.
\newblock Learning to communicate with deep multi-agent reinforcement learning.
\newblock \emph{NeurIPS}, 2016.

\bibitem[Foerster et~al.(2018)Foerster, Farquhar, Afouras, Nardelli, and Whiteson]{foerster2018counterfactual}
Jakob Foerster, Gregory Farquhar, Triantafyllos Afouras, Nantas Nardelli, and Shimon Whiteson.
\newblock Counterfactual multi-agent policy gradients.
\newblock In \emph{AAAI}, 2018.

\bibitem[Foerster et~al.(2017)Foerster, Chen, Al-Shedivat, Whiteson, Abbeel, and Mordatch]{foerster2017learning}
Jakob~N Foerster, Richard~Y Chen, Maruan Al-Shedivat, Shimon Whiteson, Pieter Abbeel, and Igor Mordatch.
\newblock Learning with opponent-learning awareness.
\newblock \emph{arXiv preprint arXiv:1709.04326}, 2017.

\bibitem[Hu et~al.(2022)Hu, Chen, Wu, Li, Yan, and Tao]{hu2022st}
Shengchao Hu, Li~Chen, Penghao Wu, Hongyang Li, Junchi Yan, and Dacheng Tao.
\newblock St-p3: End-to-end vision-based autonomous driving via spatial-temporal feature learning.
\newblock In \emph{ECCV}, 2022.

\bibitem[Hu et~al.(2023)Hu, Shen, Zhang, and Tao]{hu2023graph}
Shengchao Hu, Li~Shen, Ya~Zhang, and Dacheng Tao.
\newblock Graph decision transformer.
\newblock \emph{arXiv preprint arXiv:2303.03747}, 2023.

\bibitem[Hu et~al.(2021)Hu, Zhu, Chang, and Liang]{hu2021updet}
Siyi Hu, Fengda Zhu, Xiaojun Chang, and Xiaodan Liang.
\newblock Updet: Universal multi-agent reinforcement learning via policy decoupling with transformers.
\newblock \emph{arXiv preprint arXiv:2101.08001}, 2021.

\bibitem[Jang et~al.(2016)Jang, Gu, and Poole]{jang2016categorical}
Eric Jang, Shixiang Gu, and Ben Poole.
\newblock Categorical reparameterization with gumbel-softmax.
\newblock \emph{arXiv preprint arXiv:1611.01144}, 2016.

\bibitem[Jiang \& Lu(2018)Jiang and Lu]{jiang2018learning}
Jiechuan Jiang and Zongqing Lu.
\newblock Learning attentional communication for multi-agent cooperation.
\newblock \emph{NeurIPS}, 2018.

\bibitem[Kim et~al.(2019)Kim, Moon, Hostallero, Kang, Lee, Son, and Yi]{kim2019learning}
Daewoo Kim, Sangwoo Moon, David Hostallero, Wan~Ju Kang, Taeyoung Lee, Kyunghwan Son, and Yung Yi.
\newblock Learning to schedule communication in multi-agent reinforcement learning.
\newblock \emph{arXiv preprint arXiv:1902.01554}, 2019.

\bibitem[Kortvelesy \& Prorok(2022)Kortvelesy and Prorok]{kortvelesy2022qgnn}
Ryan Kortvelesy and Amanda Prorok.
\newblock Qgnn: Value function factorisation with graph neural networks.
\newblock \emph{arXiv preprint arXiv:2205.13005}, 2022.

\bibitem[Kuba et~al.(2022)Kuba, Chen, Wen, Wen, Sun, Wang, and Yang]{kuba2021trust}
Jakub~Grudzien Kuba, Ruiqing Chen, Munning Wen, Ying Wen, Fanglei Sun, Jun Wang, and Yaodong Yang.
\newblock Trust region policy optimisation in multi-agent reinforcement learning.
\newblock \emph{ICLR}, 2022.

\bibitem[Kurach et~al.(2020)Kurach, Raichuk, Sta{\'n}czyk, Zaj{\k{a}}c, Bachem, Espeholt, Riquelme, Vincent, Michalski, Bousquet, et~al.]{kurach2020google}
Karol Kurach, Anton Raichuk, Piotr Sta{\'n}czyk, Micha{\l} Zaj{\k{a}}c, Olivier Bachem, Lasse Espeholt, Carlos Riquelme, Damien Vincent, Marcin Michalski, Olivier Bousquet, et~al.
\newblock Google research football: A novel reinforcement learning environment.
\newblock In \emph{Proceedings of the AAAI conference on artificial intelligence}, volume~34, pp.\  4501--4510, 2020.

\bibitem[Lajoie et~al.(2021)Lajoie, Ramtoula, Wu, and Beltrame]{lajoie2021towards}
Pierre-Yves Lajoie, Benjamin Ramtoula, Fang Wu, and Giovanni Beltrame.
\newblock Towards collaborative simultaneous localization and mapping: a survey of the current research landscape.
\newblock \emph{arXiv preprint arXiv:2108.08325}, 2021.

\bibitem[Lanctot et~al.(2017)Lanctot, Zambaldi, Gruslys, Lazaridou, Tuyls, P{\'e}rolat, Silver, and Graepel]{lanctot2017unified}
Marc Lanctot, Vinicius Zambaldi, Audrunas Gruslys, Angeliki Lazaridou, Karl Tuyls, Julien P{\'e}rolat, David Silver, and Thore Graepel.
\newblock A unified game-theoretic approach to multiagent reinforcement learning.
\newblock \emph{NeurIPS}, 2017.

\bibitem[Li et~al.(2019)Li, Qin, Jiao, Yang, Wang, Wang, Wu, and Ye]{li2019efficient}
Minne Li, Zhiwei Qin, Yan Jiao, Yaodong Yang, Jun Wang, Chenxi Wang, Guobin Wu, and Jieping Ye.
\newblock Efficient ridesharing order dispatching with mean field multi-agent reinforcement learning.
\newblock In \emph{WWW}, 2019.

\bibitem[Littman(1994)]{littman1994markov}
Michael~L Littman.
\newblock Markov games as a framework for multi-agent reinforcement learning.
\newblock In \emph{Machine learning proceedings 1994}. Elsevier, 1994.

\bibitem[Liu et~al.(2021)Liu, Liu, Stone, Garg, Zhu, and Anandkumar]{liu2021coach}
Bo~Liu, Qiang Liu, Peter Stone, Animesh Garg, Yuke Zhu, and Anima Anandkumar.
\newblock Coach-player multi-agent reinforcement learning for dynamic team composition.
\newblock In \emph{ICML}, 2021.

\bibitem[Liu et~al.(2020)Liu, Wang, Hu, Hao, Chen, and Gao]{liu2020multi}
Yong Liu, Weixun Wang, Yujing Hu, Jianye Hao, Xingguo Chen, and Yang Gao.
\newblock Multi-agent game abstraction via graph attention neural network.
\newblock In \emph{AAAI}, 2020.

\bibitem[Liu et~al.(2022)Liu, Dou, Li, Xu, and Liu]{liu2022temporal}
Yuntao Liu, Yong Dou, Yuan Li, Xinhai Xu, and Donghong Liu.
\newblock Temporal dynamic weighted graph convolution for multi-agent reinforcement learning.
\newblock In \emph{CogSci}, 2022.

\bibitem[Lowe et~al.(2017)Lowe, Wu, Tamar, Harb, Pieter~Abbeel, and Mordatch]{lowe2017multi}
Ryan Lowe, Yi~I Wu, Aviv Tamar, Jean Harb, OpenAI Pieter~Abbeel, and Igor Mordatch.
\newblock Multi-agent actor-critic for mixed cooperative-competitive environments.
\newblock \emph{NeurIPS}, 2017.

\bibitem[Maddison et~al.(2016)Maddison, Mnih, and Teh]{maddison2016concrete}
Chris~J Maddison, Andriy Mnih, and Yee~Whye Teh.
\newblock The concrete distribution: A continuous relaxation of discrete random variables.
\newblock \emph{arXiv preprint arXiv:1611.00712}, 2016.

\bibitem[Mnih et~al.(2015)Mnih, Kavukcuoglu, Silver, Rusu, Veness, Bellemare, Graves, Riedmiller, Fidjeland, Ostrovski, et~al.]{mnih2015human}
Volodymyr Mnih, Koray Kavukcuoglu, David Silver, Andrei~A Rusu, Joel Veness, Marc~G Bellemare, Alex Graves, Martin Riedmiller, Andreas~K Fidjeland, Georg Ostrovski, et~al.
\newblock Human-level control through deep reinforcement learning.
\newblock \emph{nature}, 2015.

\bibitem[Niu et~al.(2021)Niu, Paleja, and Gombolay]{niu2021multi}
Yaru Niu, Rohan~R Paleja, and Matthew~C Gombolay.
\newblock Multi-agent graph-attention communication and teaming.
\newblock In \emph{AAMAS}, 2021.

\bibitem[Peng et~al.(2017)Peng, Wen, Yang, Yuan, Tang, Long, and Wang]{peng2017multiagent}
Peng Peng, Ying Wen, Yaodong Yang, Quan Yuan, Zhenkun Tang, Haitao Long, and Jun Wang.
\newblock Multiagent bidirectionally-coordinated nets: Emergence of human-level coordination in learning to play starcraft combat games.
\newblock \emph{arXiv preprint arXiv:1703.10069}, 2017.

\bibitem[Pesce \& Montana(2023)Pesce and Montana]{pesce2023learning}
Emanuele Pesce and Giovanni Montana.
\newblock Learning multi-agent coordination through connectivity-driven communication.
\newblock \emph{Machine Learning}, 2023.

\bibitem[Rashid et~al.(2020)Rashid, Samvelyan, De~Witt, Farquhar, Foerster, and Whiteson]{rashid2020monotonic}
Tabish Rashid, Mikayel Samvelyan, Christian~Schroeder De~Witt, Gregory Farquhar, Jakob Foerster, and Shimon Whiteson.
\newblock Monotonic value function factorisation for deep multi-agent reinforcement learning.
\newblock \emph{JMLR}, 2020.

\bibitem[Samvelyan et~al.(2019)Samvelyan, Rashid, De~Witt, Farquhar, Nardelli, Rudner, Hung, Torr, Foerster, and Whiteson]{samvelyan2019starcraft}
Mikayel Samvelyan, Tabish Rashid, Christian~Schroeder De~Witt, Gregory Farquhar, Nantas Nardelli, Tim~GJ Rudner, Chia-Man Hung, Philip~HS Torr, Jakob Foerster, and Shimon Whiteson.
\newblock The starcraft multi-agent challenge.
\newblock \emph{arXiv preprint arXiv:1902.04043}, 2019.

\bibitem[Schulman et~al.(2015)Schulman, Moritz, Levine, Jordan, and Abbeel]{schulman2015high}
John Schulman, Philipp Moritz, Sergey Levine, Michael Jordan, and Pieter Abbeel.
\newblock High-dimensional continuous control using generalized advantage estimation.
\newblock \emph{arXiv preprint arXiv:1506.02438}, 2015.

\bibitem[Schulman et~al.(2017)Schulman, Wolski, Dhariwal, Radford, and Klimov]{schulman2017proximal}
John Schulman, Filip Wolski, Prafulla Dhariwal, Alec Radford, and Oleg Klimov.
\newblock Proximal policy optimization algorithms.
\newblock \emph{arXiv preprint arXiv:1707.06347}, 2017.

\bibitem[Seraj et~al.(2022)Seraj, Wang, Paleja, Martin, Sklar, Patel, and Gombolay]{seraj2022learning}
Esmaeil Seraj, Zheyuan Wang, Rohan Paleja, Daniel Martin, Matthew Sklar, Anirudh Patel, and Matthew Gombolay.
\newblock Learning efficient diverse communication for cooperative heterogeneous teaming.
\newblock In \emph{AAMAS}, 2022.

\bibitem[Singh et~al.(2018)Singh, Jain, and Sukhbaatar]{singh2018learning}
Amanpreet Singh, Tushar Jain, and Sainbayar Sukhbaatar.
\newblock Learning when to communicate at scale in multiagent cooperative and competitive tasks.
\newblock \emph{arXiv preprint arXiv:1812.09755}, 2018.

\bibitem[Son et~al.(2019)Son, Kim, Kang, Hostallero, and Yi]{son2019qtran}
Kyunghwan Son, Daewoo Kim, Wan~Ju Kang, David~Earl Hostallero, and Yung Yi.
\newblock Qtran: Learning to factorize with transformation for cooperative multi-agent reinforcement learning.
\newblock In \emph{ICML}, 2019.

\bibitem[Sukhbaatar et~al.(2016)Sukhbaatar, Fergus, et~al.]{sukhbaatar2016learning}
Sainbayar Sukhbaatar, Rob Fergus, et~al.
\newblock Learning multiagent communication with backpropagation.
\newblock \emph{NeurIPS}, 2016.

\bibitem[Sunehag et~al.(2017)Sunehag, Lever, Gruslys, Czarnecki, Zambaldi, Jaderberg, Lanctot, Sonnerat, Leibo, Tuyls, et~al.]{sunehag2017value}
Peter Sunehag, Guy Lever, Audrunas Gruslys, Wojciech~Marian Czarnecki, Vinicius Zambaldi, Max Jaderberg, Marc Lanctot, Nicolas Sonnerat, Joel~Z Leibo, Karl Tuyls, et~al.
\newblock Value-decomposition networks for cooperative multi-agent learning.
\newblock \emph{arXiv preprint arXiv:1706.05296}, 2017.

\bibitem[Wang et~al.(2019)Wang, Wang, Zheng, and Zhang]{wang2019learning}
Tonghan Wang, Jianhao Wang, Chongyi Zheng, and Chongjie Zhang.
\newblock Learning nearly decomposable value functions via communication minimization.
\newblock \emph{arXiv preprint arXiv:1910.05366}, 2019.

\bibitem[Wang et~al.(2021)Wang, Zhong, Xu, and Wang]{wang2021tom2c}
Yuanfei Wang, Fangwei Zhong, Jing Xu, and Yizhou Wang.
\newblock Tom2c: Target-oriented multi-agent communication and cooperation with theory of mind.
\newblock \emph{arXiv preprint arXiv:2111.09189}, 2021.

\bibitem[Wen et~al.(2022)Wen, Kuba, Lin, Zhang, Wen, Wang, and Yang]{wen2022multi}
Muning Wen, Jakub Kuba, Runji Lin, Weinan Zhang, Ying Wen, Jun Wang, and Yaodong Yang.
\newblock Multi-agent reinforcement learning is a sequence modeling problem.
\newblock \emph{NeurIPS}, 2022.

\bibitem[Xue et~al.(2022)Xue, Yuan, Zhang, and Yu]{xue2022efficient}
Di~Xue, Lei Yuan, Zongzhang Zhang, and Yang Yu.
\newblock Efficient multi-agent communication via shapley message value.
\newblock In \emph{IJCAI}, 2022.

\bibitem[Yang \& Wang(2020)Yang and Wang]{yang2020overview}
Yaodong Yang and Jun Wang.
\newblock An overview of multi-agent reinforcement learning from game theoretical perspective.
\newblock \emph{arXiv preprint arXiv:2011.00583}, 2020.

\bibitem[Yang et~al.(2018)Yang, Luo, Li, Zhou, Zhang, and Wang]{yang2018mean}
Yaodong Yang, Rui Luo, Minne Li, Ming Zhou, Weinan Zhang, and Jun Wang.
\newblock Mean field multi-agent reinforcement learning.
\newblock In \emph{ICML}, 2018.

\bibitem[Yu et~al.(2022{\natexlab{a}})Yu, Velu, Vinitsky, Gao, Wang, Bayen, and Wu]{mappo}
Chao Yu, Akash Velu, Eugene Vinitsky, Jiaxuan Gao, Yu~Wang, Alexandre Bayen, and Yi~Wu.
\newblock The surprising effectiveness of ppo in cooperative multi-agent games.
\newblock \emph{NeurIPS}, 2022{\natexlab{a}}.

\bibitem[Yu et~al.(2022{\natexlab{b}})Yu, Velu, Vinitsky, Gao, Wang, Bayen, and Wu]{yu2022surprising}
Chao Yu, Akash Velu, Eugene Vinitsky, Jiaxuan Gao, Yu~Wang, Alexandre Bayen, and Yi~Wu.
\newblock The surprising effectiveness of ppo in cooperative multi-agent games.
\newblock \emph{NeurIPS}, 2022{\natexlab{b}}.

\bibitem[Zhou et~al.(2020)Zhou, Luo, Villella, Yang, Rusu, Miao, Zhang, Alban, Fadakar, Chen, et~al.]{zhou2020smarts}
Ming Zhou, Jun Luo, Julian Villella, Yaodong Yang, David Rusu, Jiayu Miao, Weinan Zhang, Montgomery Alban, Iman Fadakar, Zheng Chen, et~al.
\newblock Smarts: Scalable multi-agent reinforcement learning training school for autonomous driving.
\newblock \emph{arXiv preprint arXiv:2010.09776}, 2020.

\bibitem[Zhou et~al.(2023)Zhou, Wan, Wang, Wen, Wu, Wen, Yang, Yu, Wang, and Zhang]{zhou2023malib}
Ming Zhou, Ziyu Wan, Hanjing Wang, Muning Wen, Runzhe Wu, Ying Wen, Yaodong Yang, Yong Yu, Jun Wang, and Weinan Zhang.
\newblock Malib: A parallel framework for population-based multi-agent reinforcement learning.
\newblock \emph{J. Mach. Learn. Res.}, 2023.

\end{thebibliography}
\bibliographystyle{iclr2024_conference}

\clearpage
\appendices

\section{Detailed description of Environments.}
\label{sec:detail_env}

During the main experiments, we compare our method within four environments, including Predator-Prey (PP) \citep{singh2018learning}, Predator-Capture-Prey (PCP) \citep{seraj2022learning}, StarCraftII Multi-Agent Challenge (SMAC) \citep{samvelyan2019starcraft}, and Google Research Football(GRF) \citep{kurach2020google}.

\begin{itemize}[leftmargin=*]
    \item \textbf{PP}.  
    The goal is for $N$ predator agents with limited vision to find a stationary prey and move to its location. The agents in this domain all belong to the same class (i.e., identical state, observation and action spaces).

    \item \textbf{PCP}.
     We have two classes of predator and capture agents. Agents of the predator class have the goal of finding the prey with limited vision (similar to agents in PP). Agents of the capture class, have the goal of locating the prey and capturing it with an additional capture-prey action in their action-space, while not having any observation inputs (e.g., lack of scanning sensors).

    \item \textbf{SMAC}. 
    In these experiments, CommFormer controls a group of agents tasked with defeating enemy units controlled by the built-in AI. 
    The level of combat difficulty can be adjusted by varying the unit types and the number of units on both sides. 
    We measure the winning rate and compare it with state-of-the-art baseline approaches.
    Notably, the maps \textit{1o10b\_vs\_1r} and \textit{1o2r\_vs\_4r} present formidable challenges attributed to limited observational scope, while the map \textit{5z\_vs\_1ul} necessitates heightened levels of coordination to attain successful outcomes.

    \item \textbf{GRF}.
    We evaluate algorithms in the football \textit{academy scenario 3 vs. 2}, where we have 3 attackers vs. 1 defender, and 1 goalie. The three offending agents are controlled by the MARL algorithm, and the two defending agents are controlled by a built-in AI. We find that utilizing a 3 vs. 2 scenario challenges the robustness of MARL algorithms to stochasticity and sparse rewards.
\end{itemize}

\section{Detailed baselines}
\label{sec:baselines}
We compare our CommFormer with strong baselines without communication, and popular communication methods to showcase the effectiveness.
During the SMAC environments, the baseline methods are as follows, each of them is based on the CTDE paradigm to ensure fair comparison: 
(1) \textbf{MAPPO} \citep{mappo} directly apply PPO in MARL and use one shared set of parameters for all agents, without any communication.
(2) \textbf{HAPPO} \citep{kuba2021trust} implement multi-agent trust-region learning by the sequential update scheme with a monotonic improvement guarantee.
(3) \textbf{QMIX} \citep{rashid2020monotonic} incorporates a centralized value function to facilitate decentralized decision-making and efficient coordination among agents while addressing credit assignment issues.
(4) \textbf{UPDeT} \citep{hu2021updet} decouples each agent's observations into a sequence of observation entities and uses a Transformer to match different action-observation.
(5) \textbf{MAT} \citep{wen2022multi} treats cooperative MARL as sequence modeling and adopts a fixed encoder and a fully decentralized actor for each individual agent.
%
%
(6) \textbf{SMS} \citep{xue2022efficient} calculates the Shapley Message Value to explicitly evaluate each message's value, learning an efficient communication protocol in more complex scenarios.
(7) \textbf{TarMAC} \citep{das2019tarmac} utilizes an attention mechanism to integrate messages according to their relative importance. 
(8) \textbf{NDQ} \citep{wang2019learning} aims at learning nearly decomposable Q functions via communication minimization.
(9) \textbf{MAGIC} \citep{niu2021multi} makes use of hard attention to construct a dynamic communication graph, which then combines with a graph attention neural network to process the messages.
(10) \textbf{QGNN} \citep{kortvelesy2022qgnn} introduces a value factorisation method that uses a graph neural network based model.

For other domains, we benchmark our approach against a variety of state-of-the-art communication-based MARL baselines: 
(1) \textbf{CommNet} \citep{sukhbaatar2016learning} uses continuous communication for fully cooperative tasks, where the model consists of multiple agents and the communication between them is learned alongside their policy.
(2) \textbf{I3CNet} \citep{singh2018learning} controls continuous communication with a gating mechanism and uses individualized rewards for each agent to gain better performance and scalability while fixing credit assignment issues.
(3) \textbf{GA-Comm} \citep{liu2020multi} models the relationship between agents by a complete graph and proposes a novel game abstraction mechanism based on two-stage attention network.
(4) \textbf{HetNet} \citep{seraj2022learning} learns efficient and diverse communication models for coordinating cooperative heterogeneous teams based on heterogeneous graph-attention networks.

\section{Hyper-parameter Settings}
\label{sec:hyper}
During our experiments, we maintain consistency in the implementations of baseline methods by using their official repositories, and we keep all hyperparameters unchanged from their original best-performing configurations.
Specific hyperparameters used for different algorithms and tasks can be found in Tables \ref{table:common-smac} to \ref{tab:specific_SMAC}. 
To ensure a fair comparison and validate that CommFormer achieves optimal performance without compromising sample efficiency, we adopted the same hyperparameter settings as MAT \citep{wen2022multi}.

\section{Details of Experimental Results}
\begin{figure}[!htbp]
 	\centering
 	\vspace{-5pt}
 	\begin{subfigure}{0.32\linewidth}
 		\centering
 		\includegraphics[width=1.7in]{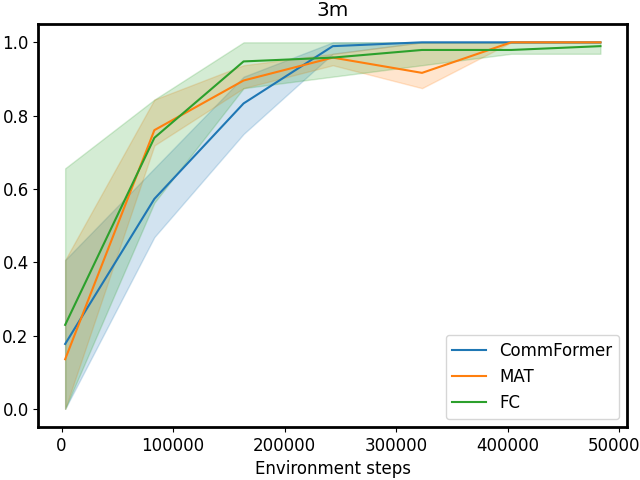} 
 	\end{subfigure}%
 	\begin{subfigure}{0.32\linewidth}
 		\centering
 		\includegraphics[width=1.7in]{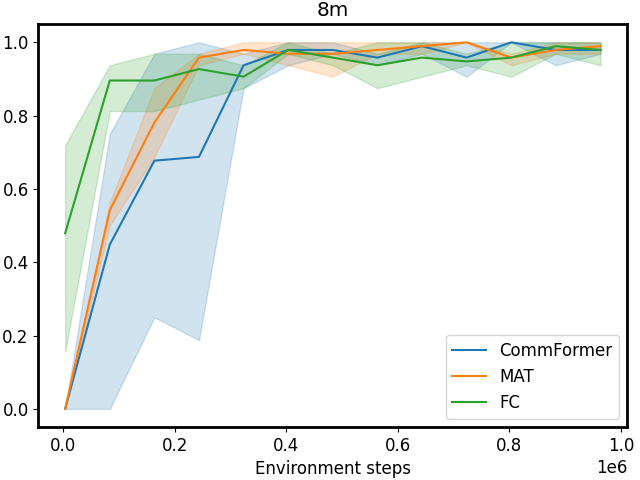} 
 	\end{subfigure}%
 	\begin{subfigure}{0.32\linewidth}
 		\centering
 		\includegraphics[width=1.7in]{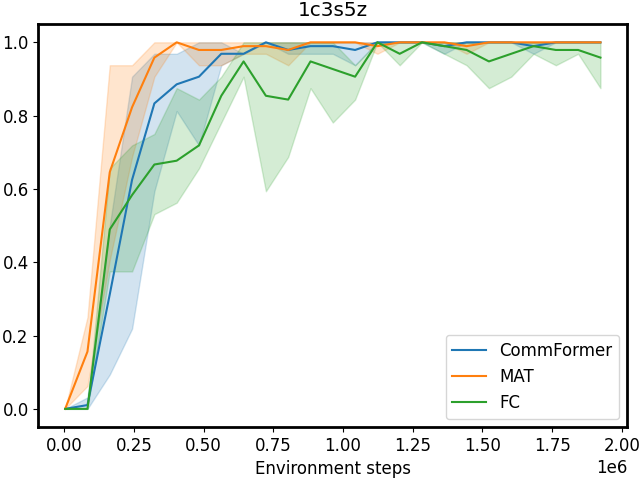} 
 	\end{subfigure}
 	\begin{subfigure}{0.32\linewidth}
 		\centering
 		\includegraphics[width=1.7in]{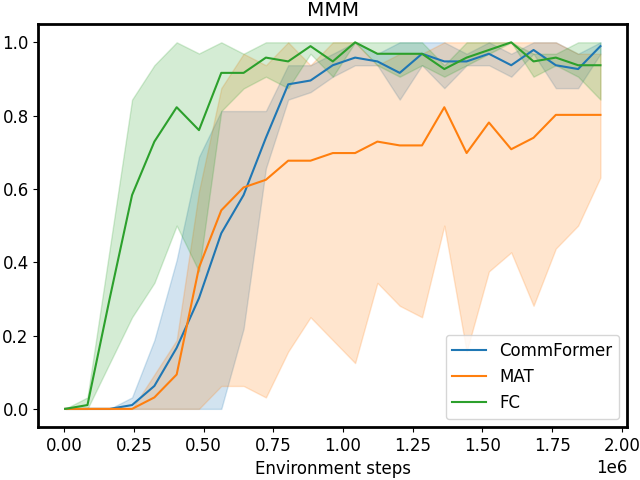}
 	\end{subfigure}%
 	\begin{subfigure}{0.32\linewidth}
 		\centering
 		\includegraphics[width=1.7in]{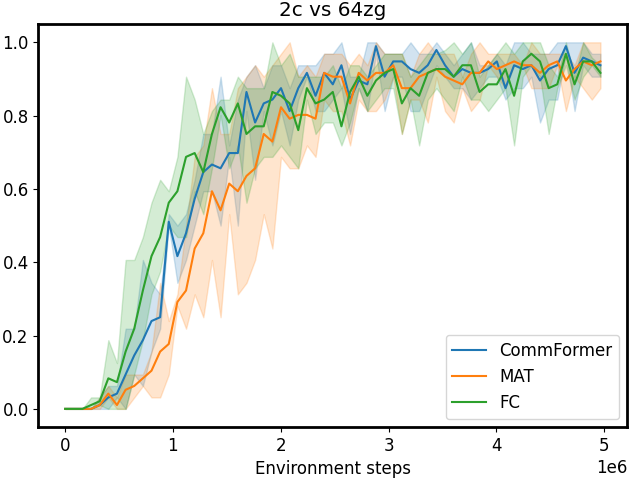}
 	\end{subfigure}%
 	\begin{subfigure}{0.32\linewidth}
 		\centering
 		\includegraphics[width=1.7in]{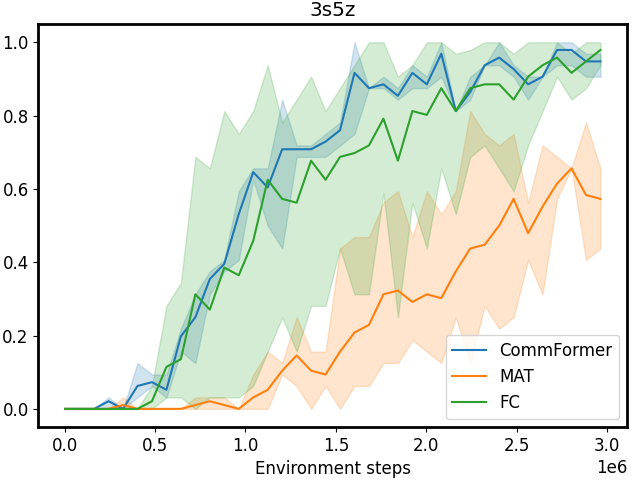}
 	\end{subfigure}
 	\begin{subfigure}{0.32\linewidth}
 		\centering
 		\includegraphics[width=1.7in]{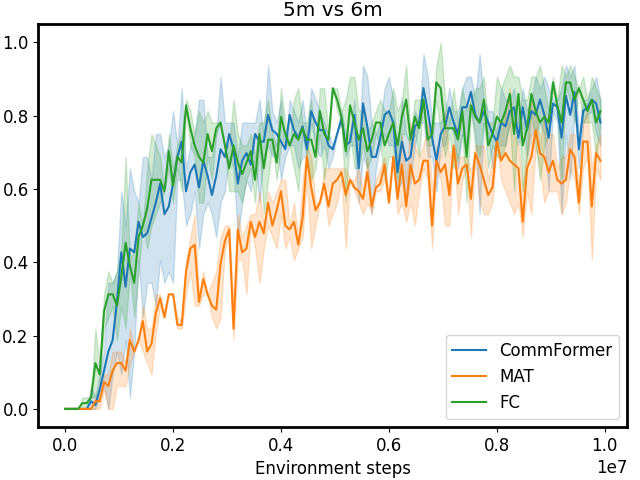}
 	\end{subfigure}%
 	\begin{subfigure}{0.32\linewidth}
 		\centering
 		\includegraphics[width=1.7in]{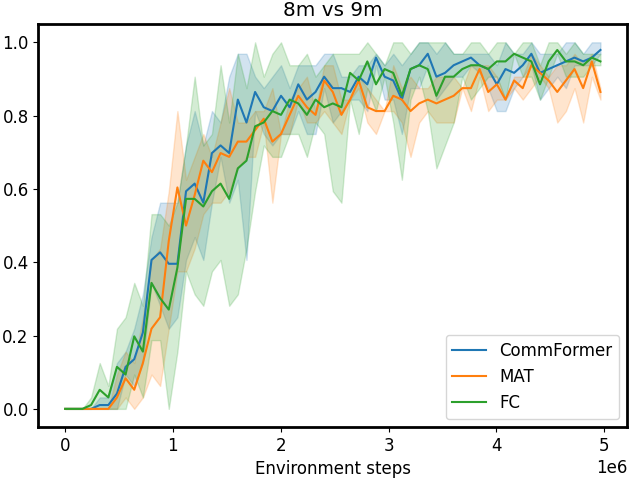}
 	\end{subfigure}%
 	\begin{subfigure}{0.32\linewidth}
 		\centering
 		\includegraphics[width=1.7in]{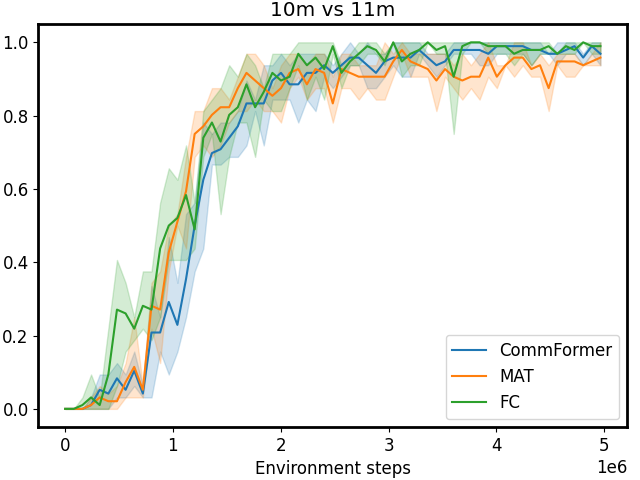}
 	\end{subfigure}
 	\begin{subfigure}{0.32\linewidth}
 		\centering
 		\includegraphics[width=1.7in]{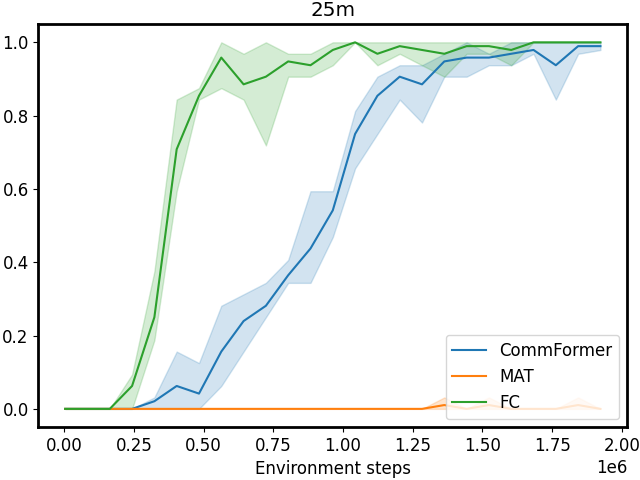} 
 	\end{subfigure}%
 	\begin{subfigure}{0.32\linewidth}
 		\centering
 		\includegraphics[width=1.7in]{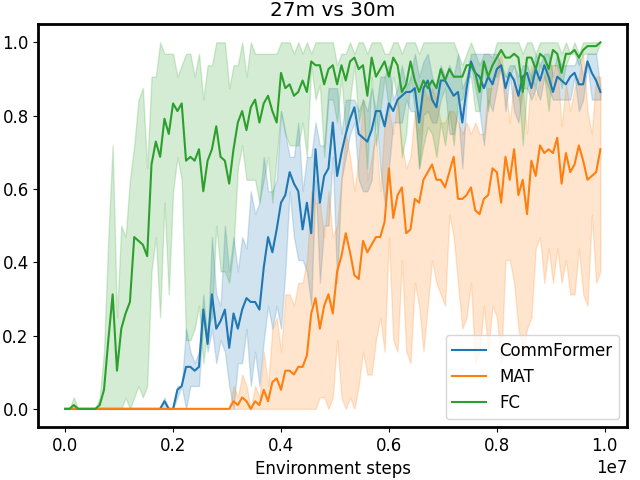}
 	\end{subfigure}%
 	\begin{subfigure}{0.32\linewidth}
 		\centering
 		\includegraphics[width=1.7in]{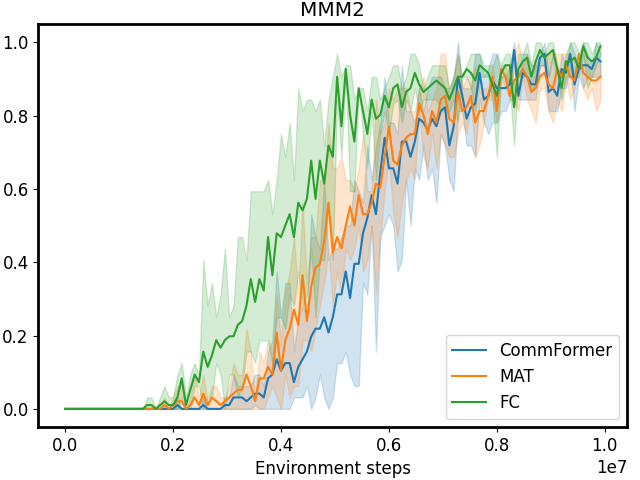} 
 	\end{subfigure}
 	\begin{subfigure}{0.45\linewidth}
 		\centering
 		\includegraphics[width=1.7in]{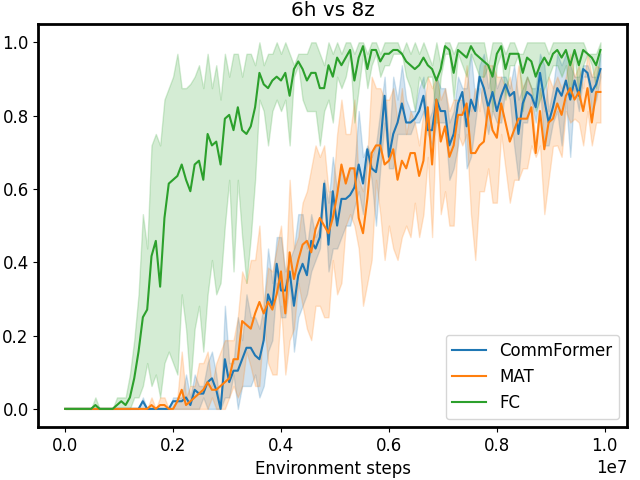}
 	\end{subfigure}%
 	\begin{subfigure}{0.45\linewidth}
 		\centering
 		\includegraphics[width=1.7in]{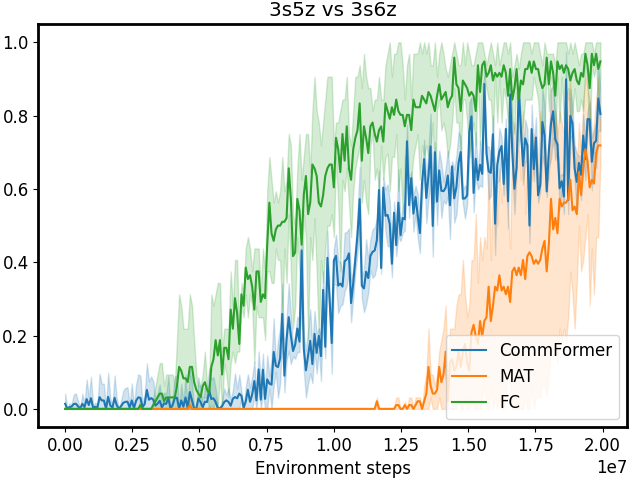} 
 	\end{subfigure}%
    \vspace{-0pt}
    \caption{\normalsize Performance comparison on SMAC tasks. CommFormer consistently outperforms strong baselines and achieves comparable performance to methods allowing information sharing among all agents, demonstrating its effectiveness regardless of variations in the number of agents. } 
    \label{fig:detail}
\end{figure}

We provide detailed training figures (Figure \ref{fig:detail}) for various methods to substantiate our claim that our approach facilitates simultaneous optimization of the communication graph and architectural parameters in an end-to-end manner, all while preserving sample efficiency.

\section{More Visual Results}
\begin{figure}[!htbp]
    \centering
    \includegraphics[width=1.0\linewidth]{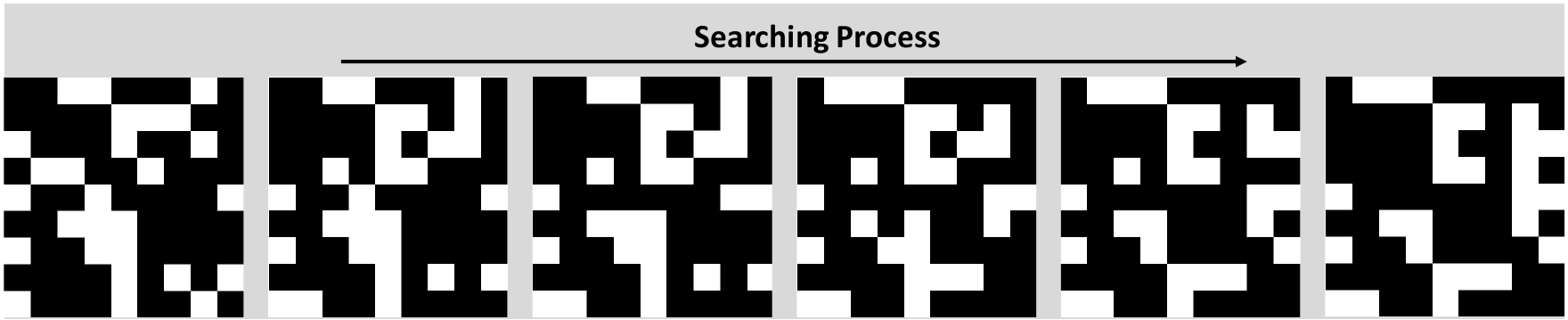}
    \caption{The searching process of CommFormer in the SMAC task \textit{1c3s5z}. In this representation, a white square corresponds to a value of 1, indicating the presence of an edge connection.  }
    \label{fig:search}
\end{figure}

We present additional visual results (Figure \ref{fig:search})  that showcase the final communication architecture obtained through our search process. 
These visualizations offer a more intuitive understanding of the architecture's evolution during training. 
As training progresses, the communication structure adapts to improve performance. 
Additionally, as we move towards the later stages of training, the model's architecture stabilizes, with only minimal changes observed, typically involving 1 or 2 edges.

\section{Application Consideration}

A possible application of this study is to create an efficient communication framework tailored for enclosed, finite environments, typical of logistics warehouses. In these settings, agent movement is limited to designated zones, and communication is facilitated through overhead wires, akin to a trolleybus system.

In contrast, open environments present unique challenges, primarily due to the potential vast distances between agents, which require wireless communication and may hinder effective communication. 
To address this, a straightforward approach could be to add bidirectional edges between agents when they come within close proximity, enabling communication between them \citep{seraj2022learning}. 
However, a more effective solution may involve a hybrid approach that considers the constraint on the available bandwidth: initially segmenting agents into groups based on proximity, followed by an internal search for an optimal communication graph within each group. 
If agent distances vary dynamically during testing, this process is repeated as necessary to adjust the communication graph in real time, ensuring continuous adaptability to changing environmental conditions.

\begin{table}[!htbp]
\caption{Common hyper-parameters used for our method in the experiments.}
 \renewcommand{\arraystretch}{1.2}
  \centering
    \begin{tabular}{cc|cc|cc}
    \toprule
    hyper-parameters & value & hyper-parameters & value & hyper-parameters & value\\
    \midrule
    critic lr & 5e-4 & actor lr & 5e-4 & use gae & True\\
    gain & 0.01 & optim eps & 1e-5 & batch size & 3200\\
    training threads & 16 & num mini-batch & 1 & rollout threads & 32\\
    entropy coef & 0.01 & max grad norm & 10 & episode length & 100\\
    optimizer & Adam & hidden layer dim & 64 & use huber loss & True\\
    \bottomrule
    \end{tabular}
    \label{table:common-smac}
\end{table}

\begin{table}[!htbp]
\caption{Specific hyper-parameters used for our method in the experiments.}
 \renewcommand{\arraystretch}{1.2}
  \centering
  \scalebox{0.9}{
    \begin{tabular}{cc|cc|cc}
    \toprule
    hyper-parameters in PP & value & hyper-parameters in PCP & value & hyper-parameters in GRF & value\\
    \midrule
    Number Agents & 3 & Number Predators & 2 & Number Agents & 3\\
    Number Enemies & 1 & Number Captures & 1 & eval episode length & 200 \\
    vision & 1 & Number Enemies & 1 & - & -\\
    eval episode length & 20 & vision & 1 & - & -\\
    - & - & eval episode length & 20 & - & -\\
    \bottomrule
    \end{tabular}}
    \label{table:common-other}
\end{table}

\begin{table}[!htbp]
\caption{Different hyper-parameters used for CommFormer in different tasks.}
 \renewcommand{\arraystretch}{1.2}
  \centering
  \label{table2}
  \resizebox{1.0\textwidth}{!}{
    \begin{tabular}{c|ccccccc}
    \toprule
    tasks & ppo epochs & ppo clip & num blocks & num heads & stacked frames & steps & $\gamma$ \\
    \midrule
    3m & 15 & 0.2 & 1 & 1 & 1 & 5e5 & 0.99 \\
    8m & 15 & 0.2 & 1 & 1 & 1 & 1e6 & 0.99 \\
    1c3s5z & 10 & 0.2 & 1 & 1 & 1 & 2e6 & 0.99 \\
    MMM & 15 & 0.2 & 1 & 1 & 1 & 2e6 & 0.99 \\
    2c vs 64zg & 10 & 0.05 & 1 & 1 & 1 & 5e6 & 0.99  \\
    3s vs 5z & 15 & 0.05 & 1 & 1 & 4 & 5e6 & 0.99  \\ 
    3s5z & 10 & 0.05 & 1 & 1 & 1 & 3e6 & 0.99  \\
    5m vs 6m & 10 & 0.05 & 1 & 1 & 1 & 1e7 & 0.99  \\
    8m vs 9m & 10 & 0.05 & 1 & 1 & 1 & 5e6 & 0.99  \\
    10m vs 11m & 10 & 0.05 & 1 & 1 & 1 &  5e6 & 0.99 \\
    25m & 15 & 0.05 & 1 & 1 & 1 & 2e6 & 0.99 \\
    27m vs 30m & 5 & 0.2 & 1 & 1 & 1 & 1e7 & 0.99 \\
    MMM2 & 10 & 0.05 & 1 & 1 & 1 & 1e7 & 0.99 \\
    6h vs 8z & 15 & 0.05 & 1 & 1 & 1 & 1e7 & 0.99 \\
    3s5z vs 3s6z & 5 & 0.05 & 1 & 1 & 1 & 2e7 & 0.99 \\
    1o10b vs 1r & 10 & 0.2 & 1 & 1 & 1 & 2e7 & 0.99 \\
    1o2r vs 4r & 5 & 0.05 & 1 & 1 & 1 & 1e7 & 0.99 \\
    5z vs 1ul & 10 & 0.05 & 1 & 1 & 1 & 1e7 & 0.99 \\
    PP & 10 & 0.05 & 1 & 1 & 1 & 1e7 & 0.99 \\
    PCP & 10 & 0.05 & 1 & 1 & 1 & 1e7 & 0.99 \\
    GRF & 10 & 0.05 & 1 & 1 & 1 & 1e7 & 0.99 \\
    \bottomrule
    \end{tabular}
    }
    \label{tab:specific_SMAC}
\end{table}

\end{document}